\documentclass[conference]{IEEEtran}
\IEEEoverridecommandlockouts
\usepackage{cite}
\usepackage{amsmath,amssymb,amsfonts}
\usepackage{algorithmic}
\usepackage{graphicx}
\usepackage{textcomp}
\usepackage{hyperref}
\usepackage{braket}
\usepackage[dvipsnames,table]{xcolor}
\usepackage[margin=1in]{geometry}
\usepackage{booktabs}    
\usepackage{pifont}      
\newcommand{\cmarkGreen}{\scalebox{1.2}{\textcolor{Green}{\ding{51}}}}
\newcommand{\xmarkRed}{\scalebox{1.2}{\textcolor{Red}{\ding{55}}}}%

\usepackage{subcaption} 
\def\BibTeX{{\rm B\kern-.05em{\sc i\kern-.025em b}\kern-.08em
    T\kern-.1667em\lower.7ex\hbox{E}\kern-.125emX}}
\begin{document}

\title{Breaking Through Barren Plateaus: Reinforcement Learning Initializations for Deep Variational Quantum Circuits \thanks{The views expressed in this article are those of the authors and do not represent the views of Wells Fargo. This article is for informational purposes only. Nothing contained in this article should be construed as investment advice. Wells Fargo makes no express or implied warranties and expressly disclaims all legal, tax, and accounting implications related to this article.}}
\author{\IEEEauthorblockN{1\textsuperscript{st} Yifeng Peng}
\IEEEauthorblockA{\textit{School of Engineering and Science} \\
\textit{Stevens Institute of Technology}\\
Hoboken, USA \\
ypeng21@stevens.edu}
\and
\IEEEauthorblockN{2\textsuperscript{nd} Xinyi Li}
\IEEEauthorblockA{\textit{School of Engineering and Science} \\
\textit{Stevens Institute of Technology}\\
Hoboken, USA \\
xli215@stevens.edu}
\and
\IEEEauthorblockN{3\textsuperscript{rd} Zhemin Zhang}
\IEEEauthorblockA{\textit{ECSE Department} \\
\textit{Rensselaer Polytechnic Institute}\\
Troy, USA \\
zhangz29@rpi.edu} 
\and
\IEEEauthorblockN{4\textsuperscript{th} Samuel Yen-Chi Chen}
\IEEEauthorblockA{\textit{Wells Fargo} \\
New York, USA \\
ycchen1989@ieee.org}
\and
\IEEEauthorblockN{5\textsuperscript{th} Zhiding Liang}
\IEEEauthorblockA{\textit{Computer Science Department} \\
\textit{Rensselaer Polytechnic Institute}\\
Troy, USA \\
liangz9@rpi.edu}
\and
\IEEEauthorblockN{6\textsuperscript{th} Ying Wang}
\IEEEauthorblockA{\textit{School of Engineering and Science} \\
\textit{Stevens Institute of Technology}\\
Hoboken, USA \\
ywang6@stevens.edu}
}
\maketitle
\begin{abstract}
Variational Quantum Algorithms (VQAs) have gained prominence as a viable framework for exploiting near-term quantum devices in applications ranging from optimization and chemistry simulation to machine learning. However, the effectiveness of VQAs is often constrained by the so-called barren plateau problem, wherein gradients diminish exponentially as system size or circuit depth increases, thereby hindering training. In this work, we propose a reinforcement learning (RL)-based initialization strategy to alleviate the barren plateau issue by reshaping the initial parameter landscape to avoid regions prone to vanishing gradients. In particular, we explore several RL algorithms (Deterministic Policy Gradient, Soft Actor-Critic, and Proximal Policy Optimization, etc.) to generate the circuit parameters (treated as “actions”) that minimize the VQAs cost function before standard gradient-based optimization. By pre-training with RL in this manner, subsequent optimization using methods such as gradient descent or Adam proceeds from a more favorable initial state. Extensive numerical experiments under various noise conditions and tasks consistently demonstrate that the RL-based initialization method significantly enhances both convergence speed and final solution quality. Moreover, comparisons among different RL algorithms highlight that multiple approaches can achieve comparable performance gains, underscoring the flexibility and robustness of our method. These findings shed light on a promising avenue for integrating machine learning techniques into quantum algorithm design, offering insights into how RL-driven parameter initialization can accelerate the scalability and practical deployment of VQAs. Opening up a promising path for the research community in machine learning for quantum, especially barren plateau problems in VQAs.
\end{abstract}

\begin{IEEEkeywords}
Barren Plateau, Initialization Methods, Variational Quantum Algorithms, Reinforcement Learning, Quantum Neural Networks
\end{IEEEkeywords}

\section{Introduction}\label{sec:introduction}

Variational Quantum Algorithms (VQAs) \cite{cerezo2021variational,bharti2022noisy} have recently emerged as a powerful paradigm for leveraging near-term quantum devices in diverse domains such as chemistry, optimization, and machine learning~\cite{schuld2020circuit,mitarai2018quantum,chen2022quantumLSTM,di2022dawn,stein2023applying,li2023pqlm,chen2022quantumCNN,yun2023quantum,lockwood2020reinforcement,skolik2022quantum,jerbi2021parametrized,chen2020variational,peng2024hyq2,peng2024qrng,peng2024qsco,peng2024quantum, peng2024hybrid, peng2025quantum}. By coupling classical optimization routines with parameterized quantum circuits (PQCs), VQAs iteratively refine a cost function --- often defined as the expectation value of a Hamiltonian or an objective operator --- to seek an optimal set of quantum circuit parameters. Despite their promise, the training of VQAs can be severely impeded by the \emph{Barren Plateau} (BP) phenomenon, where gradients vanish exponentially with system size or circuit depth. This issue has garnered significant attention, as it poses a fundamental barrier to the scalability and performance of VQAs~\cite{mcclean2018barren,cerezo2021cost,pesah2021absence,larocca2024review}.

\begin{figure*}[ht]
\centering
\includegraphics[width=0.8\textwidth]{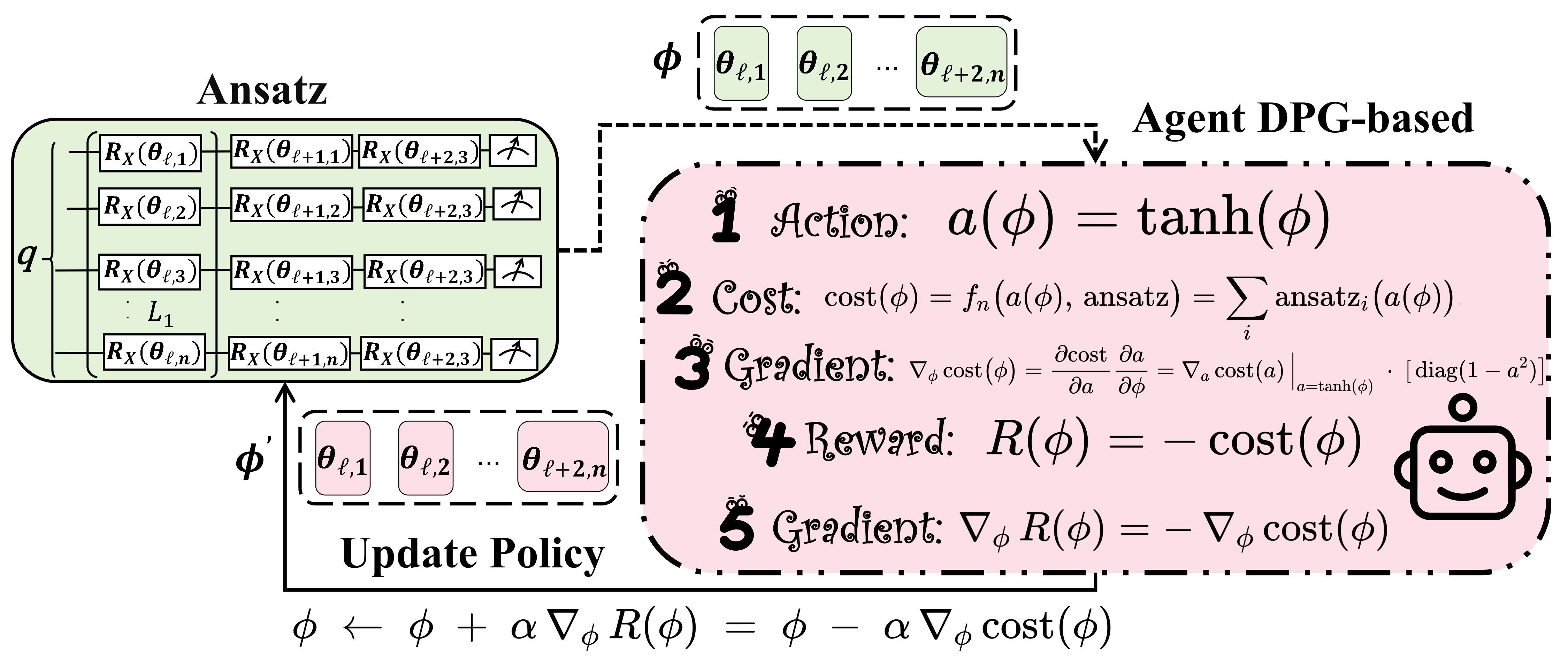} 
\caption{Training process of DPG-based RL initialization methods in Heisenberg model Ansatz.}
\label{BPlandscape}
\vspace{-5mm}
\end{figure*}

Reinforcement Learning (RL), particularly methods rooted in policy gradients, has demonstrated remarkable effectiveness in tackling high-dimensional, sequential decision-making tasks~\cite{sutton1998reinforcement,silver2014deterministic,wen2022multi,fujimoto2021minimalist}. 
In contrast, evolutionary algorithms often require a large number of population iterations and tend to converge slowly in high-dimensional searches, whereas reinforcement learning can exploit higher sampling efficiency and gradient-based updates to more rapidly establish a stronger starting point in VQA parameter initialization~\cite{salimans2017evolution}. Beyond Deterministic Policy Gradient (DPG)~\cite{DPG}, a variety of RL algorithms exist for continuous or discrete control, including Trust Region Policy Optimization (TRPO)~\cite{TRPO}, Asynchronous Advantage Actor-Critic (A3C)~\cite{A3C}, Proximal Policy Optimization (PPO)~\cite{PPO}, Deep Deterministic Policy Gradient (DDPG) \cite{DDPG} and Soft Actor-Critic (SAC)~\cite{SAC}. These methods exhibit different characteristics in terms of on-policy vs.\ off-policy updates, stochastic vs.\ deterministic policies, trust-region enforcement, sample efficiency, and time complexity. Motivated by the success of these methods in learning robust strategies under uncertainty, researchers are increasingly exploring how RL can help enhance the development of quantum computing algorithms. One of the most promising application is mitigating the challenges associated with the BP problem in VQAs.

In this work, we present a unified RL-based initialization strategy for PQCs that leverages multiple policy-gradient methods. The main idea is to treat the circuit parameters as the ``actions'' of an RL policy, training this policy to minimize the VQA cost function \emph{prior} to conventional gradient-based optimization. By thus shaping the initial parameter landscape, our approach aims to sidestep regions in parameter space where gradients are likely to vanish, thereby expediting subsequent training. 

Our key contributions are summarized as follows:
\begin{enumerate}
    \item \textbf{General RL-Based Initialization:} We propose an RL-driven framework that can accommodate a variety of policy-gradient algorithms (DDPG, TRPO, A3C, PPO, DPG, SAC) for parameter initialization in VQAs, ensuring adaptability across different use cases, circuit architectures and optimization requirements.

    \item \textbf{Avoidance of Barren Plateaus:} By pre-training circuit parameters via RL, we bypass regions with vanishing gradients, enhancing both convergence speed and the quality of the final solution.

    \item \textbf{Integration with Gradient-Based Optimizers:} We seamlessly combine our RL-based initialization with standard optimizers such as gradient descent and Adam, demonstrating consistent improvements over naive initialization strategies.

    \item \textbf{Empirical Validation and Scalability:} Through extensive numerical studies under various noise settings tasks, we show that our approach is robust and scalable, offering a promising avenue for deploying VQAs on quantum systems.
\end{enumerate}

Taken together, our findings highlight the potential of RL as a powerful tool to improve the trainability of VQAs, offering fresh perspectives on how machine learning techniques can address fundamental limitations in quantum algorithm design.

\section{Background and Related Work}\label{sec:related_work}
In this section, we survey existing research on mitigating BPs in VQA, with a particular emphasis on initialization strategies and related insights.

\paragraph{Small-Angle and Data-Informed Initialization.}
Because randomly initialized deep circuits tend to sample high-entropy regions of Hilbert space~\cite{mcclean2018barren}, \emph{small-angle} initialization schemes have been proposed to keep parameters close to identity-like transformations~\cite{grant2019initialization}. By restricting parameter values to small magnitudes, such methods aim to avoid the rapid amplitude spread that typically leads to exponentially suppressed gradients. Beyond purely random but constrained approaches, \emph{data-informed} initialization has also been explored, wherein partial classical solutions or domain-specific symmetries guide the initial parameter setting~\cite{cerezo2021cost}. Moreover, \emph{Gaussian} sampling procedures, as demonstrated by Zhang \emph{et al.}~\cite{zhang2022escaping}, systematically reduce plateau effects in deeper circuits, often outperforming zero and uniform baselines (\textit{Gaussian, zero, and uniform initialization methods are the baseline of experiments in our paper}). Taken together, these methods seek to lessen over-randomization at the outset, thereby placing the circuit closer to the manifold of interest.

\paragraph{Layerwise Training and Sequential Extensions.}
A complementary tactic is \emph{layerwise} (or ``freezing-unfreezing'') training, in which the circuit is grown incrementally: newly appended layers are initialized and trained while previously optimized layers remain ``frozen''~\cite{volkoff2021large,skolik2021layerwise}. This sequential strategy can help retain gradient signals in earlier layers even as deeper layers are introduced, potentially mitigating global randomization. Although such an approach is effective in certain settings (e.g., QAOA or VQE), it requires care to ensure that freezing does not lock in suboptimal parameters that become difficult to correct later.

\paragraph{Warm-Start and Transfer Learning.}
\emph{Warm-start} or \emph{transfer learning} strategies circumvent severe random initializations by adapting parameters pre-trained on related or smaller-scale tasks~\cite{mari2020transfer,cerezo2021cost}. Their success hinges on the degree of similarity between the pretraining and target problems: large discrepancies can place the circuit in an unfavorable parameter basin or introduce novel local minima~\cite{henderson2020quanvolutional,rivera2021avoiding}.

\paragraph{Shallow Circuits and Architectural Constraints.}
Another common BP-mitigation measure is to reduce circuit depth and employ local measurements~\cite{cerezo2021cost}. Shallow architectures, such as quantum convolutional neural networks (QCNNs)~\cite{pesah2021absence}, constrain global entanglement and slow the onset of vanishing gradients. However, overly shallow ans\"atze risk losing \emph{expressive power}~\cite{mcclean2018barren,akshay2020reachability}, potentially failing to approximate more complex target states.

\paragraph{Symmetry-Preserving and Specialized Ans\"atze.}
Finally, embedding known symmetries in VQCs can help stabilize gradient behavior by restricting parameter evolution to a more tractable subspace~\cite{larocca2022diagnosing,ragone2024lie}. Nevertheless, this approach can omit critical directions in parameter space when the enforced symmetries are overly restrictive, thereby excluding the global solution manifold~\cite{rivera2021avoiding}.

\begin{table*}[ht]
    \centering
    \caption{Comparative Overview of Key Features in Common Reinforcement Learning Algorithms}
    \label{tab:rl_comparison}
    \begin{tabular}{lcccccc}
    \toprule
    \hline
    \addlinespace[2mm]
        & \textbf{DDPG} \cite{DDPG}
        & \textbf{TRPO} \cite{TRPO}
        & \textbf{A3C$^{\dagger}$}  \cite{A3C}
        & \textbf{PPO}  \cite{PPO}
        & \textbf{DPG}  \cite{DPG}
        & \textbf{SAC}  \cite{SAC} \\
    \midrule
    \textbf{On-Policy} 
        & \xmarkRed    (\cmarkGreen)  
        & \cmarkGreen   (\cmarkGreen) 
        & \cmarkGreen   (\cmarkGreen) 
        & \cmarkGreen   (\cmarkGreen) 
        & \xmarkRed    (\cmarkGreen)  
        & \xmarkRed     (\cmarkGreen) 
        \\
    \textbf{Deterministic} 
        & \cmarkGreen  (\cmarkGreen)  
        & \xmarkRed   (\cmarkGreen)   
        & \xmarkRed    (\cmarkGreen)  
        & \xmarkRed    (\cmarkGreen)  
        & \cmarkGreen   (\cmarkGreen) 
        & \xmarkRed     (\cmarkGreen) 
        \\
    \textbf{Replay Buffer} 
        & \cmarkGreen  (\xmarkRed)  
        & \xmarkRed    (\xmarkRed)  
        & \xmarkRed    (\xmarkRed)  
        & \xmarkRed    (\xmarkRed)  
        & \xmarkRed $^\circ$  (\xmarkRed) 
        & \cmarkGreen   (\xmarkRed) 
        \\
    \textbf{Trust Region}
        & \xmarkRed   (\xmarkRed)   
        & \cmarkGreen  (\cmarkGreen)  
        & \xmarkRed    (\xmarkRed)  
        & \xmarkRed$^{\ast}$ (\xmarkRed)
        & \xmarkRed    (\xmarkRed)  
        & \xmarkRed    (\xmarkRed)  
        \\
    \rowcolor{gray!35}
    \textbf{Continuous Action Space $\diamondsuit$} 
        & \cmarkGreen  (\cmarkGreen)
        & \xmarkRed  (\cmarkGreen)
        & \xmarkRed  (\cmarkGreen)
        & \xmarkRed  (\cmarkGreen)
        & \cmarkGreen  (\cmarkGreen)
        & \xmarkRed  (\cmarkGreen)
        \\
    \textbf{Number of Hyperparameters (Adapted)}
        & (3)            
        & (3)            
        & (2)            
        & (3)            
        & (2)            
        & (3)            
        \\
    \textbf{Time Complexity$^\ddagger$ (Adapted)}
        &    ($O(N)$)    
        &     ($O(N)$)   
        &     ($O(N)$)    
        &   ($O(N)$)    
        &     ($O(N)$)   
        &     ($O(N)$)   
        \\
    \hline
    \bottomrule
    \end{tabular}
    
    \vspace{1em}
    \footnotesize
 \textbf{\textit{ Parentheses $(-)$: indicates the adapted versions in the initialization task for the VQA parameters in this paper}}.\\
    $^{\ast}$ \textbf{PPO} uses a clipped objective that approximates a trust region but is less strict than TRPO's hard KL constraint. \\
        $\diamondsuit$:  \xmarkRed~in this row indicates \textbf{default} versions are discrete but can be adapted with single-step initialization and $\tanh(\phi)$ \cite{schulman2015trust, SAC}. \\
    $^\circ$ In most recent implementations, 
    a replay buffer is employed to improve the sample efficiency of \textbf{DPG}.\\
    $^\ddagger$ \textbf{Time Complexity:} 
   Only A3C is $O(N)$ and others are  $O(N^2)$ in default versions. \\
   $^\dagger$ \textbf{A3C:} The adapted version of A3C is actually A2C due to 1 actor.\\

\end{table*}

\section{RL Initialization Methods}
\label{sec:init_methods}
In this section, we introduce the detailed derivation of several reinforcement learning-based initialization methods and summarize their differences in Table. \ref{tab:rl_comparison}. 

Table~\ref{tab:rl_comparison} provides an overview of commonly used RL algorithms applied to our initialization framework. Each algorithm offers distinct design features and operational trade-offs, such as on-policy vs. off-policy updates, deterministic vs. stochastic policies, and the use of trust-region constraints or replay buffers. For example, TRPO and PPO impose varying degrees of trust-region control, while off-policy approaches like DDPG, DPG, and SAC capitalize on data reuse via replay buffers. In practice, the choice of RL algorithm should be informed by problem characteristics such as action dimensionality, sample-efficiency requirements, and available computational resources. 

In our problem setting, the VQA parameter initialization constitutes a single-step procedure over a continuous action space restricted to \([0,\,2\pi]\). Consequently, methods such as TRPO, A3C, and PPO---originally designed for discrete actions---must be adapted to handle continuous actions. Here, we adopt a $\tanh(\phi)$ and single-step initialization representation to ensure that the parameter search remains compatible with the continuous domain required by VQA \cite{schulman2015trust, SAC}. Specifically, we redefine the policy output, gradient updates, and sampling procedures so that each algorithm (initially designed for multi-step RL) now operates in a single-step, continuous-action setting suitable for VQA parameter initialization.

\subsection{DPG (single-step initialization)} We consider a variational quantum circuit $g(\boldsymbol{a})$ acting on an initial quantum state $\ket{0}$, producing the variational state $\ket{\psi(\boldsymbol{a})} \;=\; g(\boldsymbol{a}) \ket{0}\,$.

We define a \emph{cost function}
\begin{equation}
\text{cost}(\boldsymbol{a}) \;=\; f_n\bigl(\boldsymbol{a}, \,\text{ansatz}\bigr),
\end{equation}
which might be, for instance, the expectation value of a target Hamiltonian:
\begin{equation}
\text{cost}(\boldsymbol{a}) \;=\; \bra{\psi(\boldsymbol{a})}\,H\,\ket{\psi(\boldsymbol{a})}.
\end{equation}
The goal is to find an initial parameter $\boldsymbol{a}$ that yields a reasonably low cost, or equivalently, we want to minimize $\text{cost}(\boldsymbol{a})$.

We reframe this as a \emph{reward maximization} problem \cite{DPG}, by setting $\text{reward}(\boldsymbol{a}) \;=\; -\,\text{cost}(\boldsymbol{a})$.

Hence, maximizing $\text{reward}(\boldsymbol{a})$ is equivalent to minimizing $\text{cost}(\boldsymbol{a})$. 

Let the \emph{policy parameter} be $\boldsymbol{\phi} \in \mathbb{R}^d$. We define a deterministic policy
\begin{equation}
\label{eq5}
\boldsymbol{a} \;=\; \pi_{\boldsymbol{\phi}} \;=\; \tanh(\boldsymbol{\phi}),
\end{equation}
ensuring each component of $\boldsymbol{a}$ lies in $(-1,1)$ or some scaled range. Then, the \emph{reward} in terms of $\boldsymbol{\phi}$ is
\begin{equation}
\label{eqreward1}
R(\boldsymbol{\phi})
\;=\;
\text{reward}\bigl(\,\tanh(\boldsymbol{\phi})\bigr)
\;=\;
-\,\text{cost}\bigl(\,\tanh(\boldsymbol{\phi})\bigr).
\end{equation}

Our objective is to perform \emph{gradient ascent} on $R(\boldsymbol{\phi})$. Hence we need
\begin{equation}
\nabla_{\boldsymbol{\phi}}\; R(\boldsymbol{\phi})
\;=\;
\nabla_{\boldsymbol{\phi}}
\Bigl[
-\,\text{cost}\bigl(\,\tanh(\boldsymbol{\phi})\bigr)
\Bigr].
\end{equation}
Applying the chain rule:
\begin{equation}
\label{eq:gradient}
\begin{aligned}
\nabla_{\boldsymbol{\phi}}\, R(\boldsymbol{\phi})
&=
-\,\nabla_{\boldsymbol{\phi}}\;\text{cost}\bigl(\tanh(\boldsymbol{\phi})\bigr)
\\[6pt]
&=
-\,\frac{\partial \text{cost}}{\partial \boldsymbol{a}}
\Bigl|_{\boldsymbol{a} = \tanh(\boldsymbol{\phi})}
\;\times\;
\frac{\partial \tanh(\boldsymbol{\phi})}{\partial \boldsymbol{\phi}}.
\end{aligned}
\end{equation}
Concretely, if $\boldsymbol{a} = \tanh(\boldsymbol{\phi})$, then 
\begin{equation}
\frac{\partial a_i}{\partial \phi_i}
\;=\;
1 - \bigl(a_i\bigr)^2
\quad (\text{elementwise}).
\end{equation}
Hence,
\begin{equation}
\nabla_{\phi_i}\,R(\boldsymbol{\phi})
\;=\;
-\,\sum_{j}
\frac{\partial \,\text{cost}(\boldsymbol{a})}{\partial a_j}
\Bigl(1 - a_j^2\Bigr)\,\delta_{ij},
\end{equation}
where $a_j = \tanh(\phi_j)$.

The DPG update iterates:
\begin{equation}
\boldsymbol{\phi}
\;\leftarrow\;
\boldsymbol{\phi}
\;+\;
\alpha \,\nabla_{\boldsymbol{\phi}}\,R(\boldsymbol{\phi}),
\end{equation}
where $\alpha>0$ is the learning rate (step size). After $E$ episodes of updates, we obtain the final $\boldsymbol{\phi}^{(\ast)}$. The \emph{action} $\boldsymbol{a}^{(\ast)}$ for initialization is then
\begin{equation}
\label{eq1a}
\boldsymbol{a}^{(\ast)} \;=\; \tanh\bigl(\boldsymbol{\phi}^{(\ast)}\bigr).
\end{equation}
This $\boldsymbol{a}^{(\ast)}$ serves as the \emph{initial parameter} for the quantum circuit. A precise proof of convergence for a single-step or small-step variant of DPG typically leverages:

\begin{itemize}
    \item The Lipschitz continuity of cost function $\text{cost}(\boldsymbol{a})$. 
    \item Smoothness of the $\tanh(\cdot)$ mapping. 
    \item Boundedness of the gradient. 
\end{itemize}

In short, if $\text{cost}(\boldsymbol{a})$ is continuously differentiable and its gradient is bounded, then the gradient ascent steps in $\boldsymbol{\phi}$ space will \emph{either} converge to a stationary point of $R(\boldsymbol{\phi})$ or keep making progress until limited by the environment's noise or partial observability. Because $R(\boldsymbol{\phi})=-\text{cost}(\tanh(\boldsymbol{\phi}))$ is also bounded above (assuming the cost is bounded below), the iteration
\begin{equation}
\phi \leftarrow \phi \,+\, \alpha \,\nabla_{\phi}\,R(\phi),
\end{equation}
cannot diverge indefinitely, and thus a stationary point is typically found. In practice, one must tune $\alpha$ to balance the speed of improvement and stability.

Therefore, after a finite number of episodes, we get a final $\boldsymbol{\phi}^{(\ast)}$ giving a high reward (low cost). The extracted $\boldsymbol{a}^{(\ast)} = \tanh(\boldsymbol{\phi}^{(\ast)})$ is then used as the \emph{initial parameter} for a standard VQA (or gradient-descent / Adam-based) training procedure as shown in Fig. \ref{BPlandscape}.

\subsection{DDPG (single-step initialization)}
\label{subsec:ddpg_derivation}
We provide a brief derivation of DDPG update rules used to obtain the initialization parameters \cite{DDPG}. The \emph{actor} is parameterized by a vector \(\boldsymbol{\phi} \in \mathbb{R}^d\), producing the action (circuit parameters) following Eq. (\ref{eq5}) as
\begin{equation}\label{eq:ddpg_action}
  \boldsymbol{a} \;=\; \tanh\bigl(\boldsymbol{\phi}\bigr).
\end{equation}

The \emph{critic} is a function \(Q^\theta(\boldsymbol{a})\) with parameters \(\theta\); in our simplified setting, \(Q^\theta(\boldsymbol{a})\) is modeled by a small feedforward neural network:
\begin{equation}\label{eq:critic_definition}
  Q^\theta(\boldsymbol{a}) 
  \;=\; 
  \mathbf{w}_2^\top \,\tanh\!\Bigl(\mathbf{W}_1\,\boldsymbol{a} + \mathbf{b}_1\Bigr) \;+\; b_2,
\end{equation}
where \(\theta = (\mathbf{W}_1,\mathbf{b}_1,\mathbf{w}_2,b_2)\).

The environment provides a \emph{reward} as Eq. (\ref{eqreward1}),
\begin{equation}\label{eqreward2}
  r \;=\; -\,\text{cost}\bigl(\boldsymbol{a}\bigr).
\end{equation}
So that maximizing \(r\) is equivalent to minimizing \(\text{cost}\). The critic is trained to minimize
\begin{equation}\label{eq:critic_loss}
  \mathcal{L}(\theta)
  \;=\;
  \bigl(Q^\theta(\boldsymbol{a}) \;-\; r\bigr)^2,
\end{equation}
using gradient \emph{descent}. For example, the gradient with respect to \(\mathbf{w}_2\) is
\begin{equation}\label{eq:grad_w2}
  \frac{\partial \,\mathcal{L}}{\partial \,\mathbf{w}_2}
  \;=\;
  2\,\bigl(Q^\theta(\boldsymbol{a}) - r\bigr)\,\tanh\!\Bigl(\mathbf{W}_1\,\boldsymbol{a} + \mathbf{b}_1\Bigr),
\end{equation}
and the update takes the form
\begin{equation}\label{eq:critic_update}
  \theta 
  \;\leftarrow\; 
  \theta 
  \;-\; 
  \eta_\mathrm{critic}\,\nabla_\theta\,\mathcal{L}(\theta).
\end{equation}

Meanwhile, the \emph{actor} aims to \emph{maximize} \(Q^\theta\bigl(\boldsymbol{a}\bigr)\). Since \(\boldsymbol{a} = \tanh\bigl(\boldsymbol{\phi}\bigr)\), the chain rule gives
\begin{equation}\label{eq:actor_chainrule}
  \frac{\partial \,Q^\theta(\boldsymbol{a})}{\partial \,\boldsymbol{\phi}}
  \;=\;
  \frac{\partial \,Q^\theta(\boldsymbol{a})}{\partial \,\boldsymbol{a}}
  \;\times\;
  \bigl(1 - \tanh^2(\boldsymbol{\phi})\bigr).
\end{equation}
Hence the actor update rule (gradient \emph{ascent} on \(\boldsymbol{\phi}\)) is
\begin{equation}\label{eq:actor_update}
  \boldsymbol{\phi}
  \;\leftarrow\;
  \boldsymbol{\phi}
  \;+\;
  \eta_\mathrm{actor}
  \,\nabla_{\boldsymbol{\phi}}\,
  Q^\theta\!\Bigl(\tanh\bigl(\boldsymbol{\phi}\bigr)\Bigr).
\end{equation}

By iterating these critic and actor updates, the actor parameters \(\boldsymbol{\phi}\) converge to a region (often a local optimum) that yields a high reward \(-\,\text{cost}(\boldsymbol{a})\). Thus, the resulting is an initialization for the VQA as Eq.~\eqref{eq1a}.

\subsection{PPO (single-step initialization)}
\label{subsec:ppo_derivation}
We outline here the key elements of PPO in a concise, code-agnostic manner \cite{PPO}. Consider an actor, parameterized by \(\boldsymbol{\phi}\), that defines a stochastic policy \(\pi_{\boldsymbol{\phi}}(\boldsymbol{a}\mid\boldsymbol{s})\) generally Gaussian distribution. In our setting, \(\boldsymbol{a}\) (the action) can represent the circuit parameters, while \(\boldsymbol{s}\) is an implicit state (often omitted for simplicity). The reward following Eq. (\ref{eqreward1}, \ref{eqreward2}) is $r(\boldsymbol{a})
\;=\;
-\;\text{cost}\bigl(\boldsymbol{a}\bigr)$.

In policy gradient methods, we wish to \emph{maximize} the expected reward \(\mathbb{E}\bigl[r(\boldsymbol{a})\bigr]\). When a batch of actions \(\{\boldsymbol{a}_t\}\) has been sampled from the old policy \(\pi_{\boldsymbol{\phi_{\text{old}}}}\), we define the likelihood ratio
\begin{equation}\label{eq:ratio}
r_t(\boldsymbol{\phi})
\;=\;
\frac{\pi_{\boldsymbol{\phi}}\bigl(\boldsymbol{a}_t\bigr)}{\pi_{\boldsymbol{\phi_{\text{old}}}}\bigl(\boldsymbol{a}_t\bigr)},
\end{equation}
and the corresponding \emph{advantage} \(A_t\), which is an estimate of how much better the action \(\boldsymbol{a}_t\) is than a reference baseline. In practice, \(A_t = r(\boldsymbol{a}_t) - \hat{V}\), where \(\hat{V}\) is a learned baseline (value function).

PPO modifies the standard policy gradient objective by \emph{clipping} the ratio \(r_t(\boldsymbol{\phi})\) whenever it deviates from \(1\) by more than a small threshold \(\varepsilon > 0\). Let
\begin{equation}\label{eq:ppo_surrogates}
\begin{aligned}
\mathrm{surrogate1}_t
&=\;
r_t(\boldsymbol{\phi}) \;A_t,
\\[5pt]
\mathrm{surrogate2}_t
&=\;
\mathrm{clip}\bigl(r_t(\boldsymbol{\phi}),\,1-\varepsilon,\,1+\varepsilon\bigr)\;A_t.
\end{aligned}
\end{equation}
Then the \emph{clipped objective} of PPO is
\begin{equation}\label{eq:ppo_objective}
L^{\mathrm{PPO}}(\boldsymbol{\phi})
\;=\;
\mathbb{E}\bigl[
\min\!\bigl(\mathrm{surrogate1}_t,\;\mathrm{surrogate2}_t\bigr)
\bigr].
\end{equation}
Maximizing this objective (via gradient ascent on \(\boldsymbol{\phi}\)) discourages the new policy \(\pi_{\boldsymbol{\phi}}\) from straying too far from the old policy \(\pi_{\boldsymbol{\phi_{\text{old}}}}\), maintaining stability in learning while still improving performance.

PPO proceeds by sampling actions using \(\pi_{\boldsymbol{\phi_{\text{old}}}}\), computing the clipped objective Eq. \eqref{eq:ppo_objective}, and then performing several gradient-ascent steps to update \(\boldsymbol{\phi}\):
\begin{equation}\label{eq:ppo_actor_update}
\boldsymbol{\phi}
\;\leftarrow\;
\boldsymbol{\phi}
\;+\;
\alpha
\nabla_{\boldsymbol{\phi}}\,
L^{\mathrm{PPO}}(\boldsymbol{\phi}),
\end{equation}
where \(\alpha\) is a chosen learning rate. The parameters \(\boldsymbol{\phi_{\text{old}}}\) are then set to the new \(\boldsymbol{\phi}\), and the cycle repeats.

To reduce the variance of policy-gradient estimates, a scalar baseline (value function) \(\hat{V}\) is often maintained. After each batch, one can update \(\hat{V}\) to fit the observed rewards, e.g.\ by simple mean-square minimization:
\begin{equation}\label{eq:ppo_baseline}
\hat{V}
\;\leftarrow\;
\hat{V}
\;-\;
\beta
\sum_{t}
\nabla_{\hat{V}}\!\Bigl(\,
r_t - \hat{V}
\Bigr)^{2},
\end{equation}
where \(\beta\) is a small baseline-learning rate. The advantage estimates then become \(A_t = r_t - \hat{V}\).

After multiple batches, the final parameters \(\boldsymbol{\phi}^{(\ast)}\) define a policy that \emph{samples} actions \(\boldsymbol{a}\) with high reward (i.e.\ low cost). In practice, a single \emph{deterministic} initialization can be extracted by taking the mean of the policy, e.g.:
\begin{equation}\label{eq:ppo_final_a}
\boldsymbol{a}^{(\ast)}
\;=\;
\mu\bigl(\boldsymbol{\phi}^{(\ast)}\bigr),
\end{equation}
which might be \(\boldsymbol{a}^{(\ast)} = \tanh\bigl(\boldsymbol{\phi}^{(\ast)}\bigr)\) in a bounded-parameter scenario. This \(\boldsymbol{a}^{(\ast)}\) serves as the initialization for subsequent fine-tuning.

\begin{figure*}[htbp]
    \centering
  
    \begin{subfigure}[b]{0.24\textwidth}
        \centering
        \includegraphics[width=\textwidth]{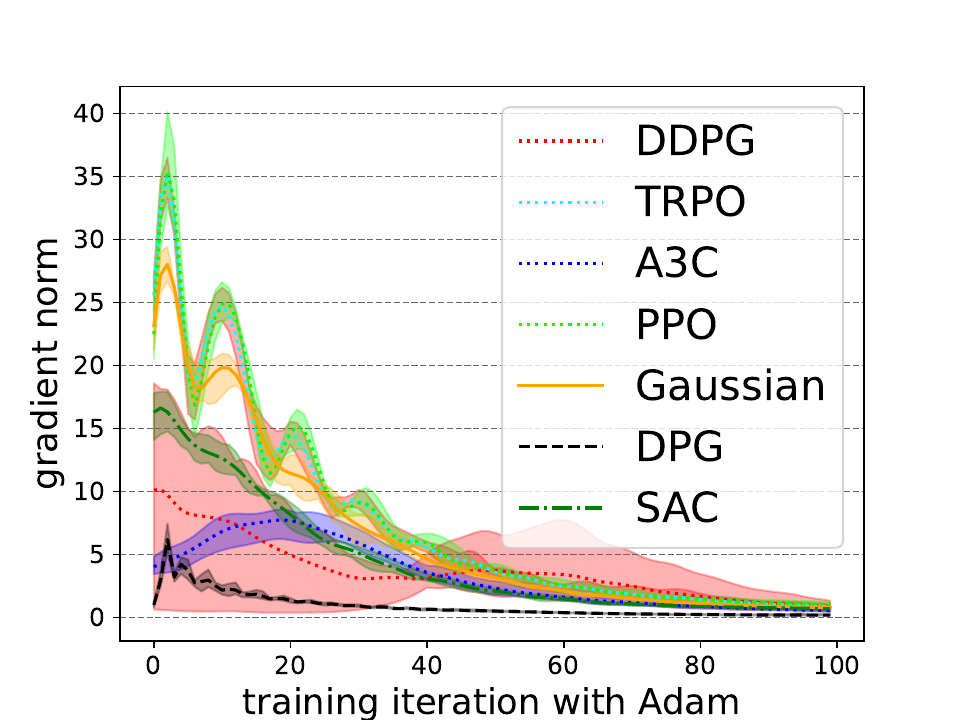}
        \caption{}
        \label{figRLbench:sub1}
    \end{subfigure}
    \hfill
    \begin{subfigure}[b]{0.24\textwidth}
        \centering
        \includegraphics[width=\textwidth]{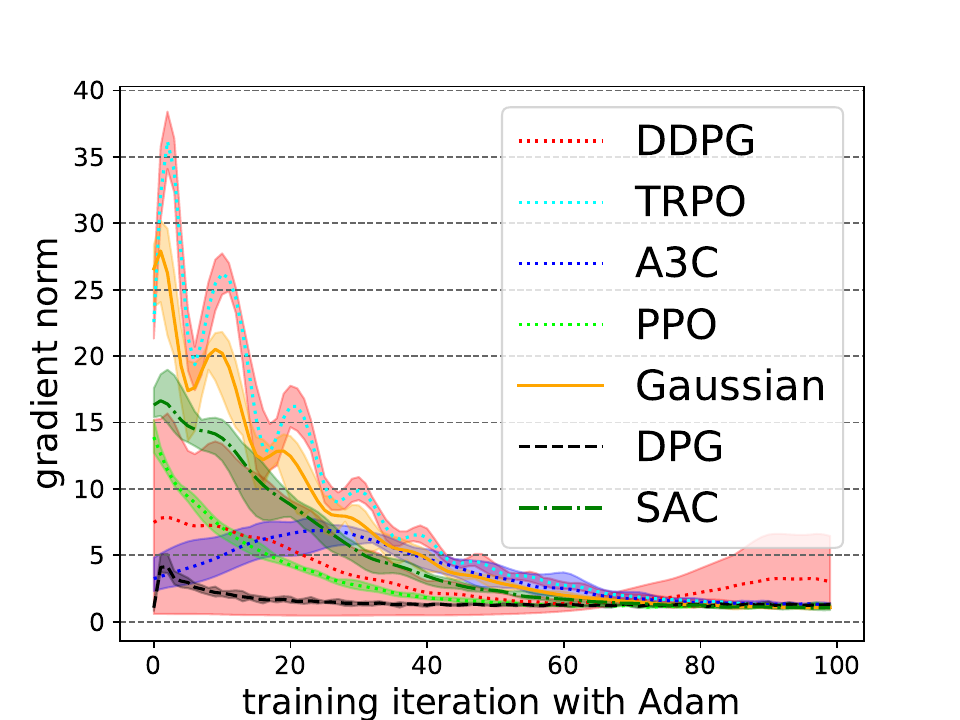}
        \caption{}
        \label{figRLbench:sub2}
    \end{subfigure}
    \hfill
    \begin{subfigure}[b]{0.24\textwidth}
        \centering
        \includegraphics[width=\textwidth]{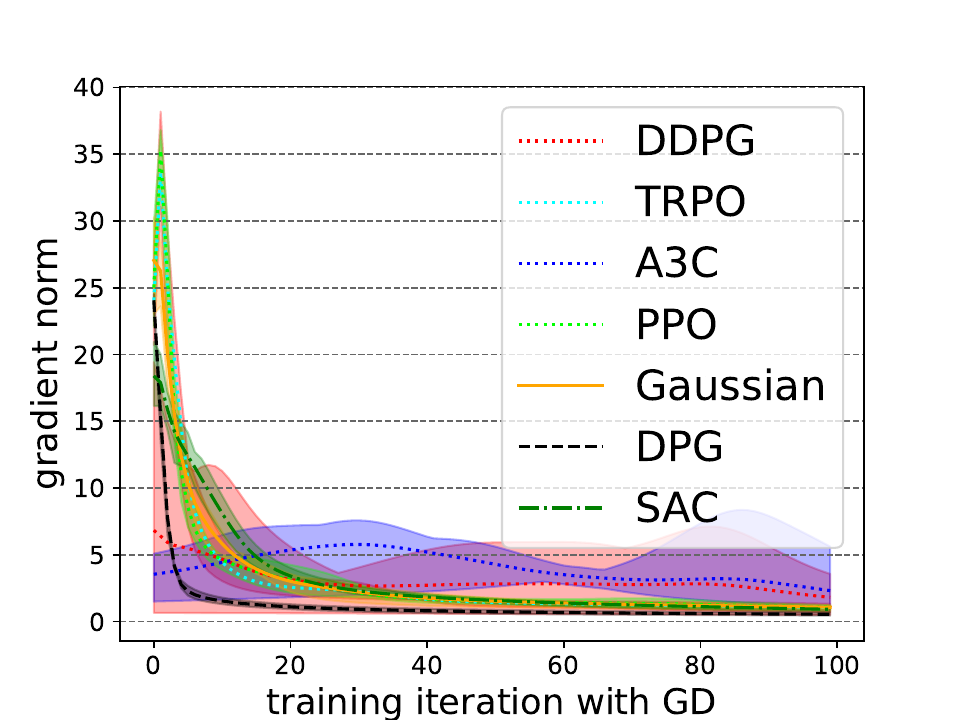}
        \caption{}
        \label{figRLbench:sub3}
    \end{subfigure}
    \hfill
    \begin{subfigure}[b]{0.24\textwidth}
        \centering
        \includegraphics[width=\textwidth]{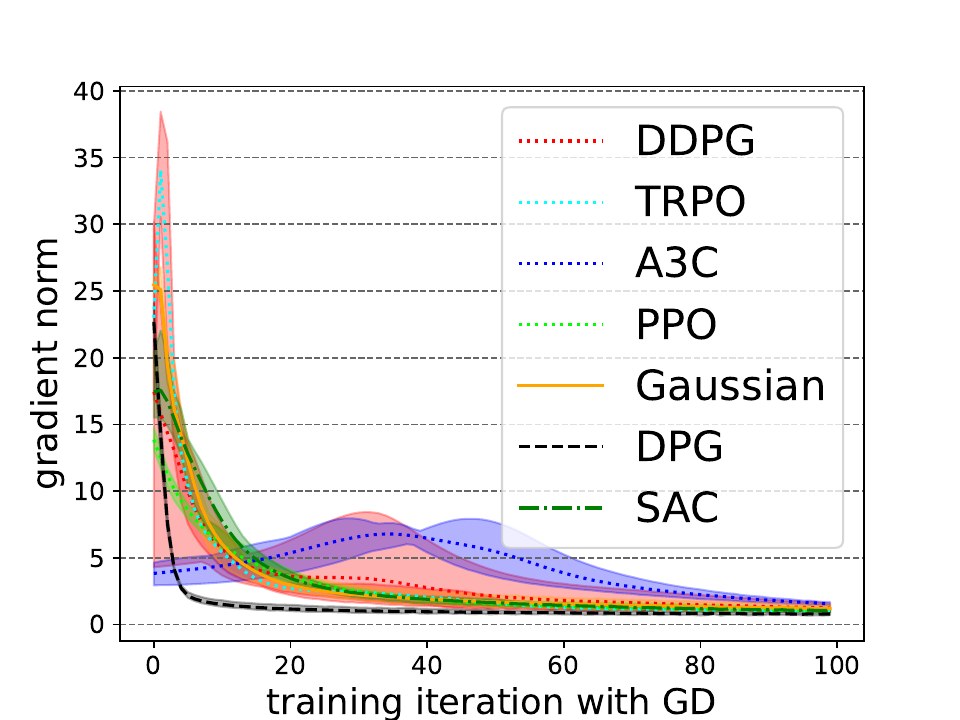}
        \caption{}
        \label{figRLbench:sub4}
    \end{subfigure}

    \vspace{1em}  

    \begin{subfigure}[b]{0.24\textwidth}
        \centering
        \includegraphics[width=\textwidth]{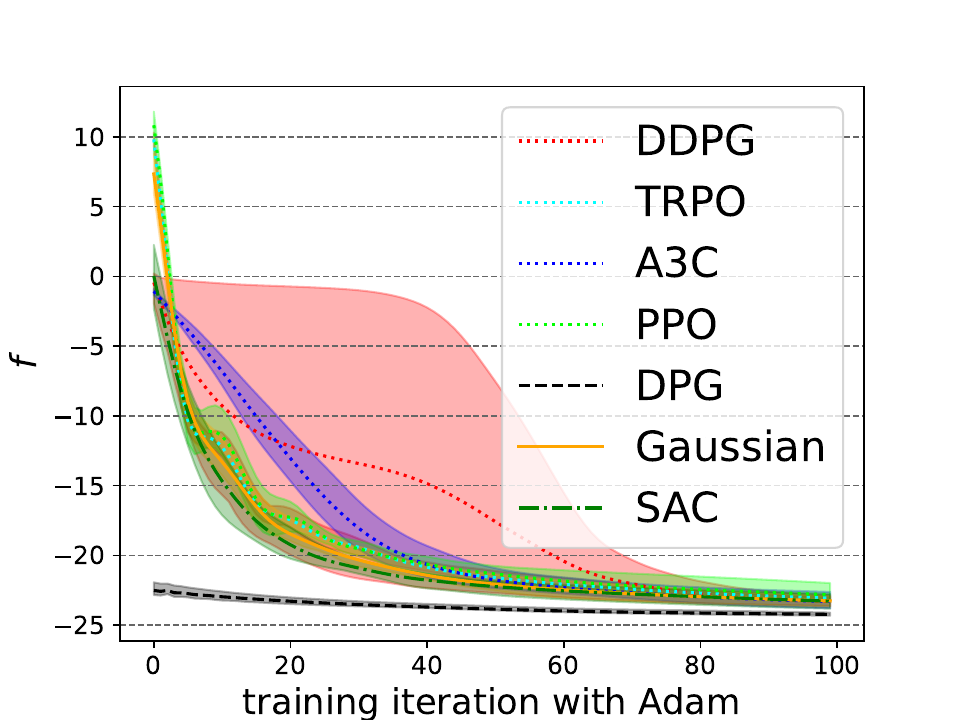}
        \caption{}
        \label{figRLbench:sub5}
    \end{subfigure}
    \hfill
    \begin{subfigure}[b]{0.24\textwidth}
        \centering
        \includegraphics[width=\textwidth]{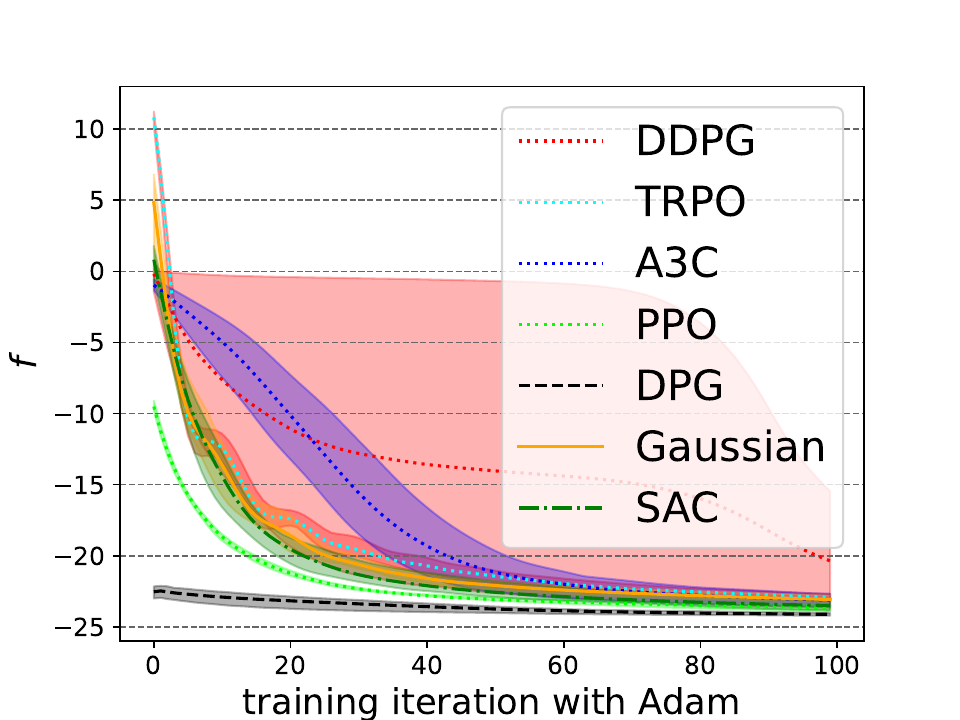}
        \caption{}
        \label{figRLbench:sub6}
    \end{subfigure}
    \hfill
    \begin{subfigure}[b]{0.24\textwidth}
        \centering
        \includegraphics[width=\textwidth]{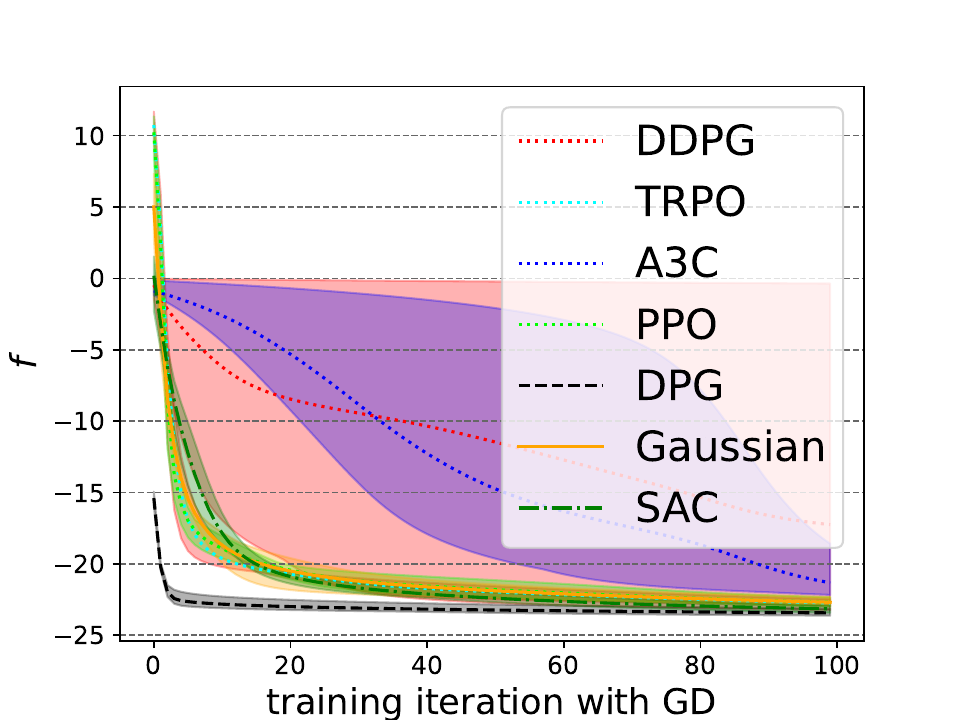}
        \caption{}
        \label{figRLbench:sub7}
    \end{subfigure}
    \hfill
    \begin{subfigure}[b]{0.24\textwidth}
        \centering
        \includegraphics[width=\textwidth]{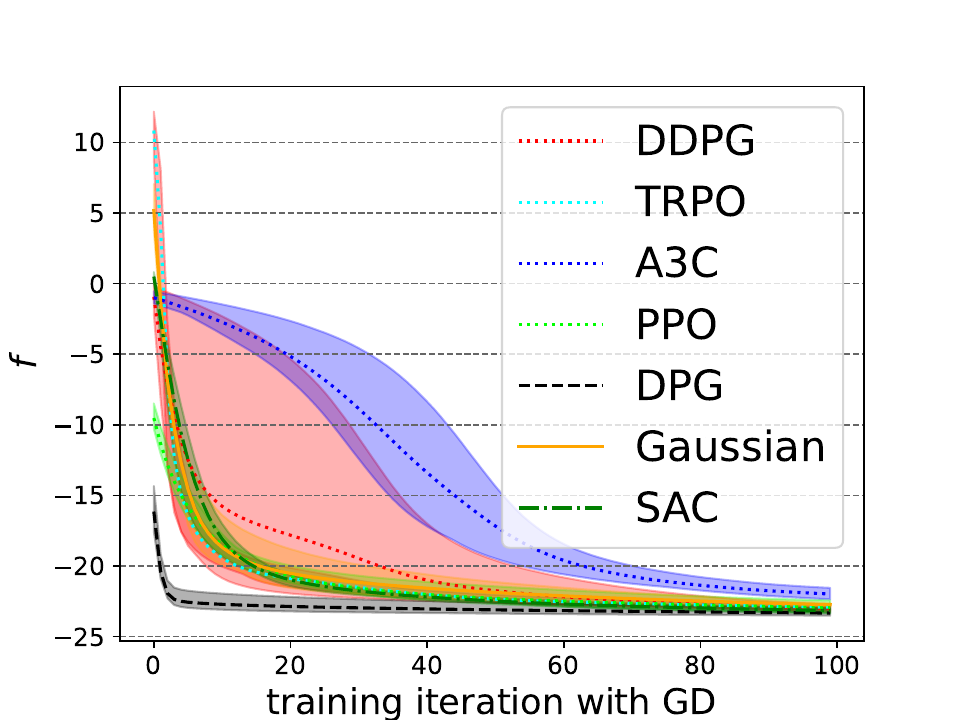}
        \caption{}
        \label{figRLbench:sub8}
    \end{subfigure}

    \caption{Numerical results of finding the ground energy of the Heisenberg model (RL Algorithms). The first row shows training results of $l_2$ norm with the Gradient Descent and Adam optimizer, where Figures \ref{figRLbench:sub1} and \ref{figRLbench:sub3} illustrate $l_2$ norm with accurate gradients. Figures \ref{figRLbench:sub2} and \ref{figRLbench:sub4}) show $l_2$ norm with noisy gradients. The second row shows the loss function as Eq. (\ref{eq:HeisenbergHamiltonian}) with the Gradient Descent and Adam optimizer, where Figures \ref{figRLbench:sub5} and \ref{figRLbench:sub7} illustrate loss with accurate gradients. Figures \ref{figRLbench:sub6} and \ref{figRLbench:sub8}) show loss with noisy gradients. Each line denotes the average of $5$ rounds of optimizations.}
    \label{fig:figRLbench}
\end{figure*}

\subsection{SAC (single-step initialization)}
\label{subsec:sac_derivation}

SAC \cite{SAC} seeks to optimize a policy by balancing high expected reward with high entropy of policy. In our setting, the \emph{action} \(\boldsymbol{a} \in \mathbb{R}^d\) represents parameters of a quantum circuit, and the \emph{reward} is $r(\boldsymbol{a})
  \;=\;
  -\;\text{cost}\bigl(\boldsymbol{a}\bigr)$.

We maintain both an \emph{actor} (policy) and a \emph{critic} (Q-function). In SAC, we define a stochastic policy with parameter vector \(\boldsymbol{\phi}\).  Let
\begin{equation}\label{eq:sac_actor_mean}
  \boldsymbol{\mu}
  \;=\;
  \tanh\!\bigl(\boldsymbol{\phi}\bigr),
\end{equation}
and sample the action by adding Gaussian noise:
\begin{equation}\label{eq:sac_actor_action}
  \boldsymbol{a}
  \;=\;
  \boldsymbol{\mu}
  \;+\;
  \sigma \,\boldsymbol{\epsilon}, 
  \quad
  \boldsymbol{\epsilon} \sim \mathcal{N}(\boldsymbol{0}, \mathbf{I}),
\end{equation}
where \(\sigma\) is a fixed scalar standard deviation.

The critic is a parameterized function \(Q^\theta(\boldsymbol{a})\) that estimates the expected return when action \(\boldsymbol{a}\) is taken. A standard objective is to regress \(Q^\theta(\boldsymbol{a})\) to match the observed reward:
\begin{equation}\label{eq:sac_critic_loss}
  \mathcal{L}_\mathrm{critic}(\theta)
  \;=\;
  \Bigl(
    Q^\theta(\boldsymbol{a})
    \;-\;
    r(\boldsymbol{a})
  \Bigr)^2.
\end{equation}
Minimizing \(\mathcal{L}_\mathrm{critic}\) via gradient descent updates the critic parameters \(\theta\).

SAC incorporates an entropy term to encourage exploration. 
Specifically, define the \emph{log-likelihood} of \(\boldsymbol{a}\) under the policy:
\begin{equation}\label{eq:sac_logpi}
  \log \pi_{\boldsymbol{\phi}}(\boldsymbol{a})
  \;=\;
  -\;\frac12
  \sum_{i=1}^d
  \Bigl(
    \frac{a_i - \mu_i}{\sigma}
  \Bigr)^2
  \;-\;
  \frac{d}{2}\,\log\!\bigl(2\pi\sigma^2\bigr),
\end{equation}
where \(\boldsymbol{\mu} = \tanh(\boldsymbol{\phi})\).  Let \(\alpha > 0\) be a temperature parameter that weights the entropy term.  The \emph{actor} aims to \emph{maximize} the expected value of
\begin{equation}\label{eq:sac_actor_objective}
  Q^\theta(\boldsymbol{a}) \;+\; \alpha \,H\bigl(\boldsymbol{a}\bigr),
  \quad
  \text{where}
  \quad
  H(\boldsymbol{a})
  \;=\;
  -\;\log \pi_{\boldsymbol{\phi}}(\boldsymbol{a}).
\end{equation}
In practice, one may define the actor’s loss to be
\begin{equation}
  \mathcal{L}_\mathrm{actor}(\boldsymbol{\phi})
  \;=\;
  -\;\Bigl(
    Q^\theta(\boldsymbol{a}) \;+\; \alpha \,\bigl[-\log \pi_{\boldsymbol{\phi}}(\boldsymbol{a})\bigr]
  \Bigr),
\end{equation}
so that \(\boldsymbol{\phi}\) is updated via \emph{gradient descent} on \(\mathcal{L}_\mathrm{actor}\).  

If \(\boldsymbol{a} = \boldsymbol{\mu} + \sigma\,\boldsymbol{\epsilon}\) with \(\boldsymbol{\mu} = \tanh(\boldsymbol{\phi})\), then the derivative \(\tfrac{\partial \boldsymbol{a}}{\partial \boldsymbol{\mu}} = \mathbf{I}\).  Meanwhile, since \(\boldsymbol{\mu} = \tanh(\boldsymbol{\phi})\),
\begin{equation}\label{eq:sac_mu_phi}
  \frac{\partial \mu_i}{\partial \phi_i}
  \;=\;
  1 - \mu_i^2.
\end{equation}
Hence, the chain rule yields the update directions for \(\boldsymbol{\phi}\) involving \(\tfrac{\partial Q^\theta}{\partial \boldsymbol{a}}\) and \(\tfrac{\partial}{\partial \boldsymbol{a}}\bigl(-\alpha\,\log\pi_{\boldsymbol{\phi}}(\boldsymbol{a})\bigr)\).  

\begin{itemize}
  \item \textbf{Critic update:} For each sampled action \(\boldsymbol{a}\), observe reward \(r\) and compute 
  \(\bigl(Q^\theta(\boldsymbol{a}) - r\bigr)^2\).  Update \(\theta\) by gradient descent on \(\mathcal{L}_\mathrm{critic}\).
  \item \textbf{Actor update:} Update \(\boldsymbol{\phi}\) to \emph{minimize} 
  \(\mathcal{L}_\mathrm{actor} = -\bigl[\,Q^\theta(\boldsymbol{a}) + \alpha\,H(\boldsymbol{a})\bigr]\), 
  where \(\boldsymbol{a}\) is sampled from the current policy \(\pi_{\boldsymbol{\phi}}\).
\end{itemize}

The same as Eq.~\eqref{eq:ppo_final_a}, the single \emph{deterministic} initialization is taking from $\boldsymbol{a}^{(\ast)}\;=\;\tanh\!\bigl(\boldsymbol{\phi}^{(\ast)}\bigr)$.

\subsection{A3C (single-step initialization) (actor = 1)}
\label{subsec:a3c_derivation}
The A3C \cite{A3C} algorithm combines a policy network (\emph{actor}) and a value function (\emph{critic}) to optimize actions for high reward. In our setting, a single-step (or single-action) version is sufficient to produce a set of parameters \(\boldsymbol{a}\in\mathbb{R}^d\) for a quantum circuit. The reward is still defined as $r(\boldsymbol{a}) 
\;=\; -\;\text{cost}\bigl(\boldsymbol{a}\bigr)$.

The actor is parameterized by \(\boldsymbol{\phi}\), which produces a mean \(\boldsymbol{\mu} = \tanh(\boldsymbol{\phi})\). The action (circuit parameters) \(\boldsymbol{a}\) is then sampled from a Gaussian distribution:
\begin{equation}
\boldsymbol{a}
\;=\;
\boldsymbol{\mu}
\;+\;
\sigma\,\boldsymbol{\epsilon},
\quad
\boldsymbol{\epsilon}
\sim
\mathcal{N}(\boldsymbol{0}, \mathbf{I}).
\end{equation}
Thus, the \emph{log-likelihood} of sampling \(\boldsymbol{a}\) under this is
\begin{equation}\label{eq:a3c_logpi}
\log \pi_{\boldsymbol{\phi}}(\boldsymbol{a})
\;=\;
-\;\frac12
\sum_{i=1}^d
\Bigl(
  \frac{a_i - \mu_i}{\sigma}
\Bigr)^2
\;-\;
\frac{d}{2}
\,\log\!\bigl(2\pi\sigma^2\bigr),
\end{equation}
where \(\boldsymbol{\mu} = \tanh(\boldsymbol{\phi})\) and \(\sigma\) is a fixed standard deviation.

The critic is a function \(V^\theta(\boldsymbol{a})\) designed to predict the expected future reward given the action \(\boldsymbol{a}\). For simplicity, we treat \(\boldsymbol{a}\) itself as the ``state,'' so
\begin{equation}
V^\theta(\boldsymbol{a})
\;\approx\;
\mathbb{E}\bigl[r(\boldsymbol{a})\bigr].
\end{equation}
We train \(V^\theta\) by minimizing a mean-square loss between the predicted value \(V^\theta(\boldsymbol{a})\) and the observed reward \(r(\boldsymbol{a})\):
\begin{equation}\label{eq:a3c_critic_loss}
\mathcal{L}_{\text{critic}}(\theta)
\;=\;
\bigl(
  V^\theta(\boldsymbol{a})
  - r(\boldsymbol{a})
\bigr)^2.
\end{equation}

In A3C, the \emph{advantage} of a sampled action is 
\begin{equation}\label{eq:a3c_advantage}
A(\boldsymbol{a})
\;=\;
r(\boldsymbol{a})
\;-\;
V^\theta\bigl(\boldsymbol{a}\bigr).
\end{equation}
The actor aims to \emph{maximize} the total reward, so it performs gradient \emph{ascent} on the log-likelihood weighted by the advantage. Namely, if \(\boldsymbol{\mu}=\tanh(\boldsymbol{\phi})\), then
\begin{equation}\label{eq:a3c_actor_update}
\nabla_{\boldsymbol{\phi}}\;\mathcal{L}_{\text{actor}}
\;=\;
-\;\bigl[
  A(\boldsymbol{a})
  \,\nabla_{\boldsymbol{\phi}}\,
  \log \pi_{\boldsymbol{\phi}}(\boldsymbol{a})
\bigr].
\end{equation}
Because 
\(\nabla_{\boldsymbol{\mu}}\,\log \pi_{\boldsymbol{\phi}}(\boldsymbol{a}) \;=\; \frac{\boldsymbol{a} - \boldsymbol{\mu}}{\sigma^2}\)
and 
\(\boldsymbol{\mu} = \tanh(\boldsymbol{\phi})\) implies 
\(\tfrac{\partial \mu_i}{\partial \phi_i} = 1 - \mu_i^2\),
the complete chain rule yields
\begin{equation}\label{eq:a3c_grad_chain}
\nabla_{\boldsymbol{\phi}}
\;\log \pi_{\boldsymbol{\phi}}(\boldsymbol{a})
\;=\;
\Bigl(\frac{\boldsymbol{a}-\boldsymbol{\mu}}{\sigma^2}\Bigr)
\;\odot\;
\bigl(1 - \boldsymbol{\mu}^2\bigr),
\end{equation}
where \(\odot\) denotes elementwise multiplication. Hence we can update \(\boldsymbol{\phi}\) in a gradient-ascent manner with learning rate \(\eta_{\text{actor}}\):
\begin{equation}\label{eq:a3c_actor_grad}
\boldsymbol{\phi}
\;\leftarrow\;
\boldsymbol{\phi}
\;+\;
\eta_{\text{actor}}
\,
A(\boldsymbol{a})
\,
\Bigl(\frac{\boldsymbol{a}-\boldsymbol{\mu}}{\sigma^2}\Bigr)
\odot
\bigl(1-\boldsymbol{\mu}^2\bigr).
\end{equation}

A3C applies the above critic and actor updates repeatedly (often in parallel across multiple agents). After sufficient training, the final parameters \(\boldsymbol{\phi}^{(\ast)}\) produce a policy that yields high reward (low cost). The same as Eq.~\eqref{eq:ppo_final_a}, the single \emph{deterministic} initialization is taking from  $\boldsymbol{a}^{(\ast)}
  \;=\;
  \tanh\!\bigl(\boldsymbol{\phi}^{(\ast)}\bigr)$.

\subsection{TRPO (single-step initialization)}
\label{subsec:trpo_derivation}

TRPO \cite{TRPO} modifies the basic policy gradient approach by enforcing a \emph{trust region} constraint on how much the policy can change between updates. Let \(\pi_{\boldsymbol{\phi}}\) be the actor (policy) with parameters \(\boldsymbol{\phi}\), generally Gaussian distribution, and define the \emph{surrogate objective}:
\begin{equation}\label{eq:trpo_surrogate}
  L(\boldsymbol{\phi};\,\boldsymbol{\phi}_{\text{old}})
  \;=\;
  \mathbb{E}_{\boldsymbol{a} \sim \pi_{\boldsymbol{\phi_{\text{old}}}}}\Bigl[
    \frac{\pi_{\boldsymbol{\phi}}(\boldsymbol{a})}{\pi_{\boldsymbol{\phi_{\text{old}}}}(\boldsymbol{a})}
    \;\cdot\;
    A(\boldsymbol{a})
  \Bigr],
\end{equation}
where \(\boldsymbol{\phi_{\text{old}}}\) is the parameter vector used to sample \(\boldsymbol{a}\), and \(A(\boldsymbol{a})\) is the advantage of \(\boldsymbol{a}\). TRPO’s core idea is to \emph{maximize} \(L(\boldsymbol{\phi};\,\boldsymbol{\phi}_{\text{old}})\) subject to a constraint on the KL divergence between \(\pi_{\boldsymbol{\phi}}\) and \(\pi_{\boldsymbol{\phi_{\text{old}}}}\):
\begin{equation}\label{eq:trpo_constraint}
  \text{maximize}\quad L(\boldsymbol{\phi})
  \quad
  \text{subject~to}
  \quad
  \mathrm{D_{KL}}\bigl(\pi_{\boldsymbol{\phi_{\text{old}}}},\,\pi_{\boldsymbol{\phi}} \bigr)
  \;\le\;
  \delta,
\end{equation}
where \(\delta > 0\) is a small trust-region threshold (e.g.\ 0.01). In practice, one typically uses the \emph{Fisher information matrix} of \(\pi_{\boldsymbol{\phi_{\text{old}}}}\) to approximate the KL divergence, yielding a \emph{natural gradient} update. We define the ratio
\begin{equation}\label{eq:trpo_ratio}
  r_t(\boldsymbol{\phi})
  \;=\;
  \frac{\pi_{\boldsymbol{\phi}}(\boldsymbol{a}_t)}{\pi_{\boldsymbol{\phi_{\text{old}}}}(\boldsymbol{a}_t)},
\end{equation}
where \(\boldsymbol{a}_t\) was sampled using \(\boldsymbol{\phi_{\text{old}}}\). The surrogate objective is then
\begin{equation}\label{eq:trpo_surrogate_restate}
  L(\boldsymbol{\phi})
  \;=\;
  \mathbb{E}\Bigl[
    r_t(\boldsymbol{\phi})\,A_t
  \Bigr].
\end{equation}
A gradient \emph{ascent} on \(L(\boldsymbol{\phi})\) alone might produce large, destabilizing policy steps.

TRPO enforces a constraint on the average KL divergence:
\begin{equation}
  \mathrm{D_{KL}}\!\bigl[
    \pi_{\boldsymbol{\phi_{\text{old}}}}(\boldsymbol{a}),
    \;\pi_{\boldsymbol{\phi}}(\boldsymbol{a})
  \bigr]
  \;\le\;
  \delta.
\end{equation}
To implement this, we approximate \(\mathrm{D_{KL}}\) to second order around \(\boldsymbol{\phi_{\text{old}}}\), forming a local quadratic model:
\begin{equation}\label{eq:trpo_quadratic_approx}
  \mathrm{D_{KL}}\!\bigl(\boldsymbol{\phi_{\text{old}}},\boldsymbol{\phi}\bigr)
  \;\approx\;
  \tfrac12\,\bigl(\boldsymbol{\phi} - \boldsymbol{\phi_{\text{old}}}\bigr)^\top
  \mathbf{F}
  \,\bigl(\boldsymbol{\phi} - \boldsymbol{\phi_{\text{old}}}\bigr),
\end{equation}
where \(\mathbf{F}\) is the Fisher information matrix evaluated at \(\boldsymbol{\phi_{\text{old}}}\). Consequently, the TRPO update amounts to solving
\begin{equation}\label{eq:trpo_optimization}
  \max_{\Delta \boldsymbol{\phi}}
  \bigl[
    \nabla_{\boldsymbol{\phi}}\,
    L(\boldsymbol{\phi_{\text{old}}})
  \bigr]
  \cdot
  \Delta \boldsymbol{\phi}
  \quad
  \text{subject~to}
  \quad
  \tfrac12\,\Delta \boldsymbol{\phi}^\top \mathbf{F}\,\Delta \boldsymbol{\phi}
  \;\le\;
  \delta.
\end{equation}
One can show the solution to Eq. (\eqref{eq:trpo_optimization}) is
\begin{equation}\label{eq:trpo_stepdir}
  \Delta \boldsymbol{\phi}
  \;=\;
  \sqrt{
    \frac{2\,\delta}
    {
      \nabla_{\boldsymbol{\phi}}\,L
      \cdot
      \mathbf{F}^{-1}
      \,\nabla_{\boldsymbol{\phi}}\,L
    }
  }
  \;\mathbf{F}^{-1}
  \nabla_{\boldsymbol{\phi}}\,L,
\end{equation}
which is essentially a \emph{scaled natural gradient} step. In practice, \(\mathbf{F}^{-1}\) is approximated using the \emph{conjugate gradient} method, and a backtracking line search ensures that the final update satisfies \(\mathrm{D_{KL}} \le \delta\) and actually improves the surrogate objective.

After multiple iterations, the final parameters \(\boldsymbol{\phi}^{(\ast)}\) define a policy with improved reward (lower cost). Similar to other methods in Eq.~\eqref{eq:ppo_final_a}, the single \emph{deterministic} initialization is taking from  $\boldsymbol{a}^{(\ast)}
  \;=\;
  \tanh\!\bigl(\boldsymbol{\phi}^{(\ast)}\bigr)$.

\begin{figure*}[htbp]
    \centering
  
    \begin{subfigure}[b]{0.24\textwidth}
        \centering
        \includegraphics[width=\textwidth]{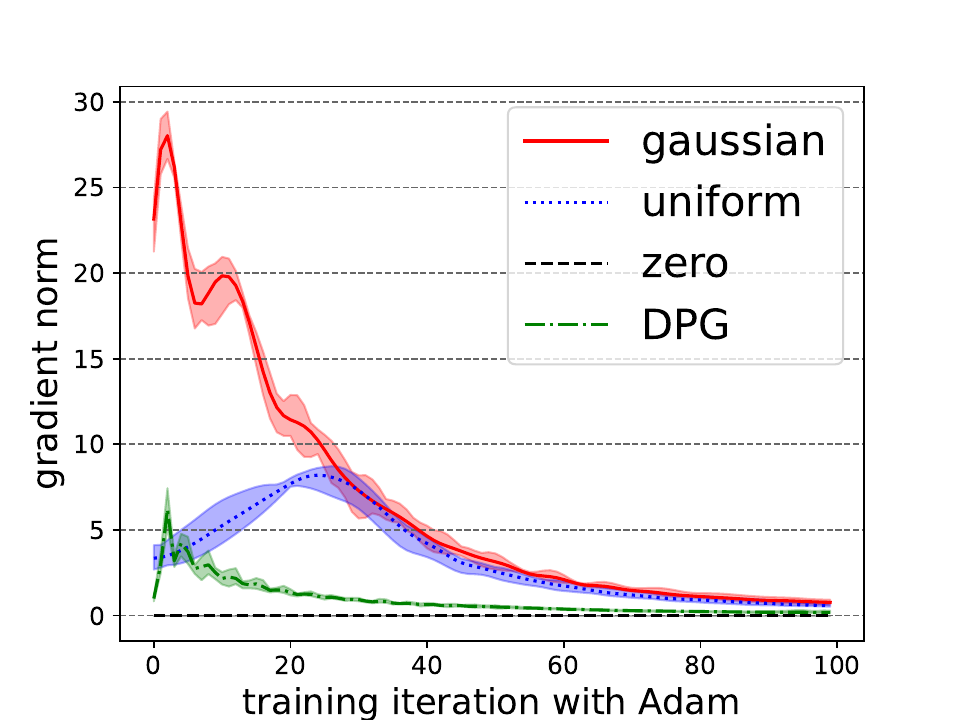}
        \caption{}
        \label{Heisenbergfig:sub1}
    \end{subfigure}
    \hfill
    \begin{subfigure}[b]{0.24\textwidth}
        \centering
        \includegraphics[width=\textwidth]{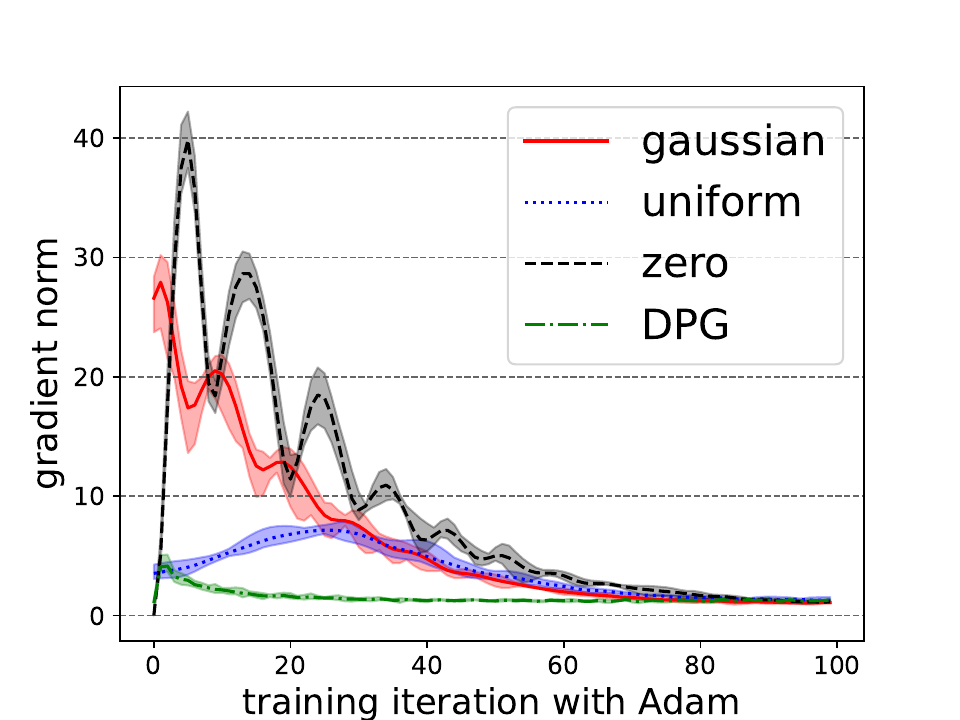}
        \caption{}
        \label{Heisenbergfig:sub2}
    \end{subfigure}
    \hfill
    \begin{subfigure}[b]{0.24\textwidth}
        \centering
        \includegraphics[width=\textwidth]{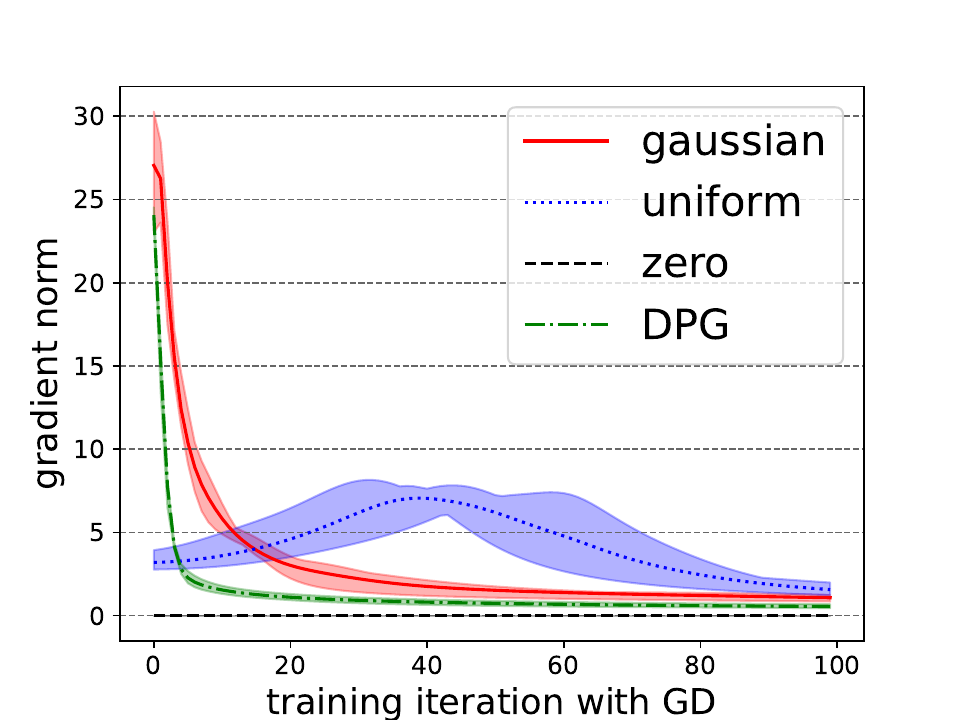}
        \caption{}
        \label{Heisenbergfig:sub3}
    \end{subfigure}
    \hfill
    \begin{subfigure}[b]{0.24\textwidth}
        \centering
        \includegraphics[width=\textwidth]{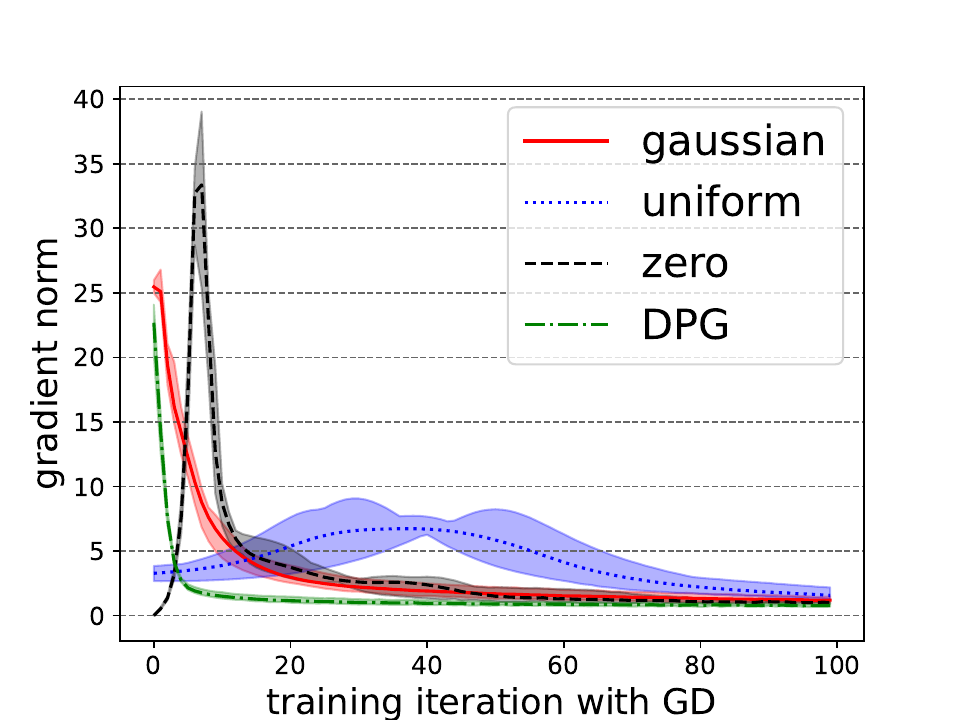}
        \caption{}
        \label{Heisenbergfig:sub4}
    \end{subfigure}

    \vspace{1em}  

    \begin{subfigure}[b]{0.24\textwidth}
        \centering
        \includegraphics[width=\textwidth]{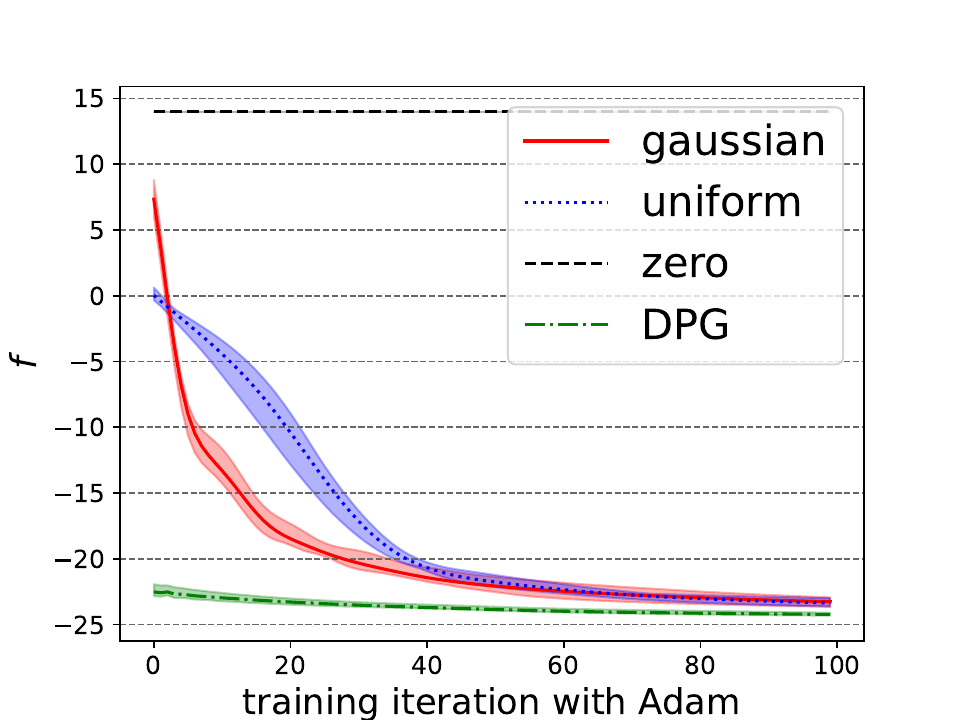}
        \caption{}
        \label{Heisenbergfig:sub5}
    \end{subfigure}
    \hfill
    \begin{subfigure}[b]{0.24\textwidth}
        \centering
        \includegraphics[width=\textwidth]{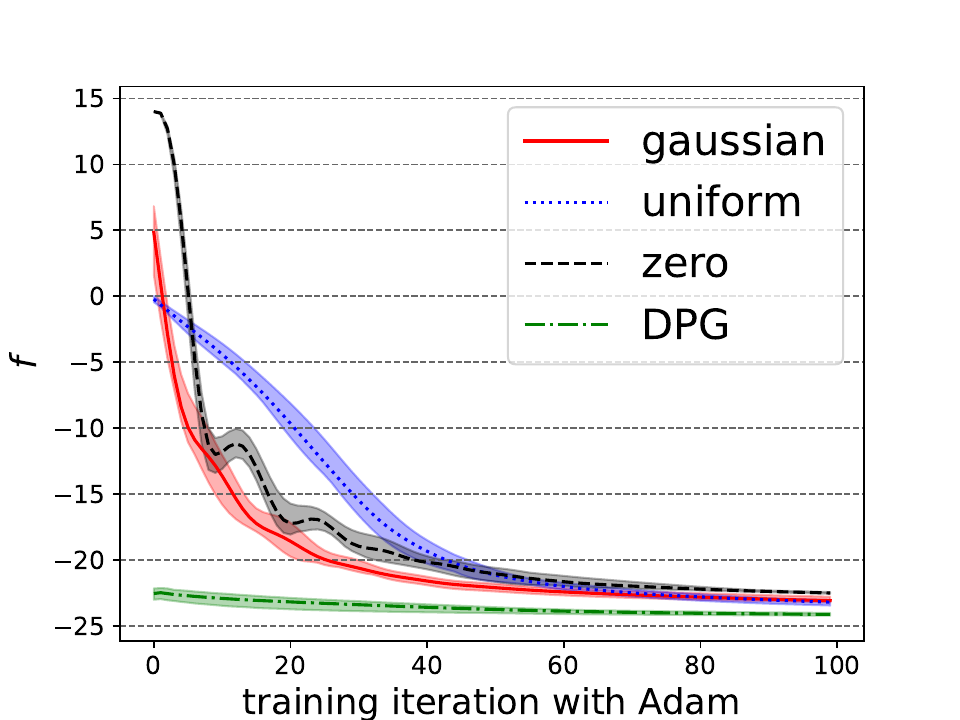}
        \caption{}
        \label{Heisenbergfig:sub6}
    \end{subfigure}
    \hfill
    \begin{subfigure}[b]{0.24\textwidth}
        \centering
        \includegraphics[width=\textwidth]{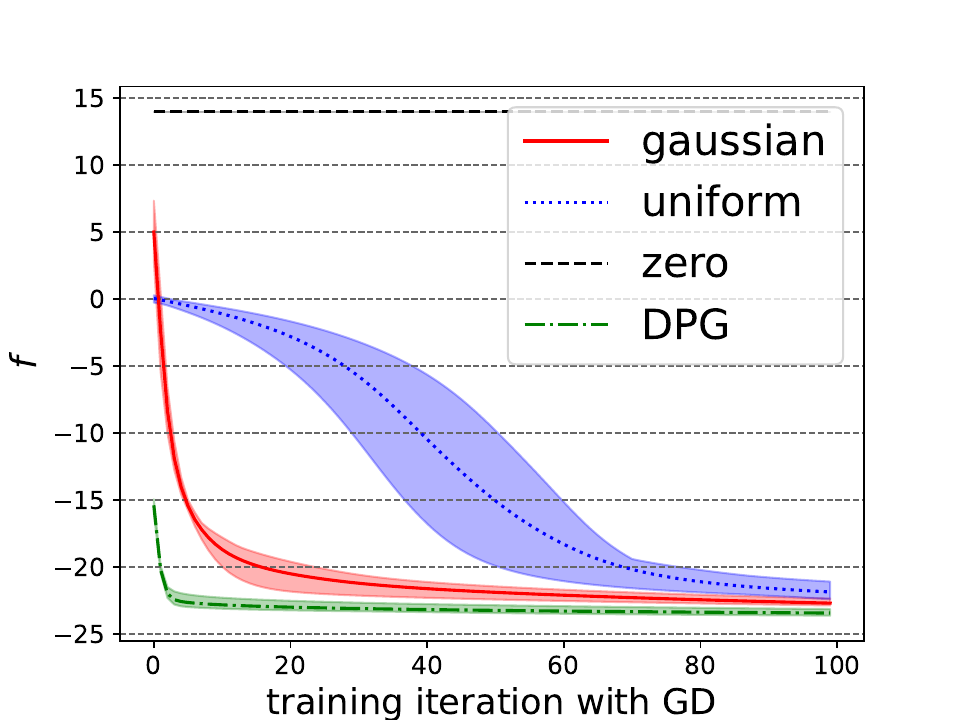}
        \caption{}
        \label{Heisenbergfig:sub7}
    \end{subfigure}
    \hfill
    \begin{subfigure}[b]{0.24\textwidth}
        \centering
        \includegraphics[width=\textwidth]{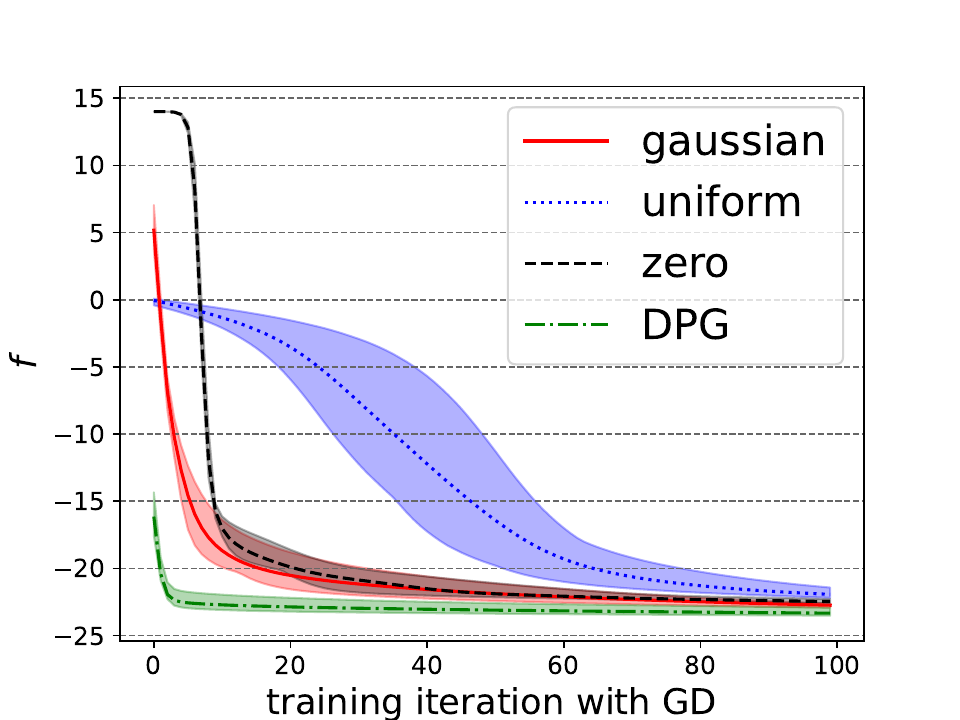}
        \caption{}
        \label{Heisenbergfig:sub8}
    \end{subfigure}

    \caption{Numerical results of finding the ground energy of the Heisenberg model (DPG). The first row
shows training results of $l_2$ norm with the Gradient Descent and Adam optimizer, where Figures \ref{Heisenbergfig:sub1} and \ref{Heisenbergfig:sub3} illustrate $l_2$ norm with accurate gradients. Figures \ref{Heisenbergfig:sub2} and \ref{Heisenbergfig:sub4}) show $l_2$ norm with noisy gradients. The second row shows the loss function as Eq. (\ref{eq:HeisenbergHamiltonian}) with the Gradient Descent and Adam optimizer, where Figures \ref{Heisenbergfig:sub5} and \ref{Heisenbergfig:sub7} illustrate loss with accurate gradients. Figures \ref{Heisenbergfig:sub6} and \ref{Heisenbergfig:sub8}) show loss with noisy gradients. Each line denotes the average of $5$ rounds of optimizations.}
    \label{fig:all_five}
    \vspace{-2mm}
\end{figure*}

\section{Experiments}
\label{sec:experiments}
\textbf{Hyperparameters Setting} We conduct all experiments using a fixed budget of \(50\) training episodes for each method. The step size \(\alpha\) is set to \(0.05\) for DPG, TRPO, A3C, and PPO, whereas both the actor and critic learning rates in DDPG are configured to \(0.02\). In the case of TRPO, the target KL divergence is fixed at \(0.01\); the clipping ratio in PPO is set to \(0.20\); and the entropy coefficient in SAC is chosen as \(0.01\).

In this section, we formalize the general framework of VQAs and introduce the corresponding notation. A typical VQA can be viewed as an optimization task involving the function
\begin{equation}
\label{eq:generalVQAf}
f(\boldsymbol{\theta}) \;=\; \mathrm{Tr}\Bigl[
\,O \, V(\boldsymbol{\theta}) \, \rho_{\mathrm{in}} \, V(\boldsymbol{\theta})^\dagger
\Bigr],
\end{equation}
where \(V(\boldsymbol{\theta})\) denotes a parameterized unitary operator, \(\rho_{\mathrm{in}}\) represents the input quantum state, and \(O\) is a specified observable. 
The output \(f(\boldsymbol{\theta})\) depends on the parameter vector \(\boldsymbol{\theta}\) and the choice of \(\rho_{\mathrm{in}}\).

\subsection{Heisenberg Model}
\label{subsec:heisenberg}

In this work, we focus on approximating the ground state and ground energy of the Heisenberg model~\cite{bonechi1992heisenberg}, whose Hamiltonian is given by
\begin{equation}
\label{eq:HeisenbergHamiltonian}
H \;=\; \sum_{i=1}^{N-1} \Bigl(X_i X_{i+1} + Y_i Y_{i+1} + Z_i Z_{i+1}\Bigr),
\end{equation}
where \(N\) is the number of qubits. 
The operators \(X_i\), \(Y_i\), and \(Z_i\) act nontrivially on the \(i\)-th qubit (as Pauli \(X\), \(Y\), or \(Z\)) and trivially on all other qubits, that is,
\begin{align}
X_i &= I^{\otimes (i-1)} \otimes X \otimes I^{\otimes (N-i)}, \\
Y_i &= I^{\otimes (i-1)} \otimes Y \otimes I^{\otimes (N-i)}, \\
Z_i &= I^{\otimes (i-1)} \otimes Z \otimes I^{\otimes (N-i)}.
\end{align}

We employ the loss function from Eq.~\eqref{eq:generalVQAf} using the initial state \(\lvert 0\rangle^{\otimes N}\) and the observable \(H\). 
Minimizing this loss function effectively drives the system toward the ground energy of \(H\).

Our ansatz consists of a circuit with \(N=15\) qubits arranged in \(L = 10\) layers of \(\mathrm{R_y}\,\mathrm{R_x}\,\mathrm{CZ}\) blocks. 
Within each block, we first apply \(\mathrm{CZ}\) gates to all neighboring qubit pairs \(\{(1,2),\dots,(N,1)\}\), followed by \(\mathrm{R_x}\) and \(\mathrm{R_y}\) rotations on every qubit. 
In total, the circuit contains \(300\) adjustable parameters.

Because each term of \(H\) involves at most \(S=2\) non-identity Pauli operators, this setting corresponds to \((S,K)=(2,18)\). 
Following the convention ~\cite{zhang2022escaping}, we set
\[
\gamma^2 \;=\; \frac{1}{4\,S\,(K+2)} \;=\; \frac{1}{160}.
\]

For training, we employ both gradient descent (GD) and Adam optimizer, each with a learning rate of \(0.01\). 

\begin{figure*}
    \centering
  
    \begin{subfigure}[b]{0.3\textwidth}
        \centering
        \includegraphics[width=\textwidth]{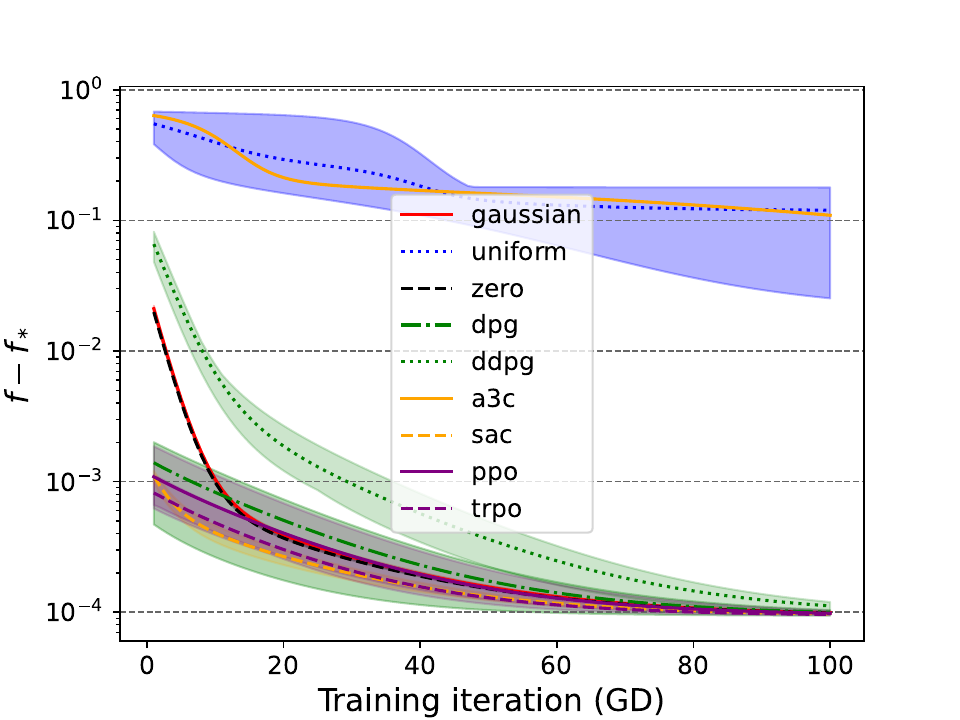}
        \caption{}
        \label{chemenergyfigrl:sub1}
    \end{subfigure}
    \hfill
    \begin{subfigure}[b]{0.3\textwidth}
        \centering
        \includegraphics[width=\textwidth]{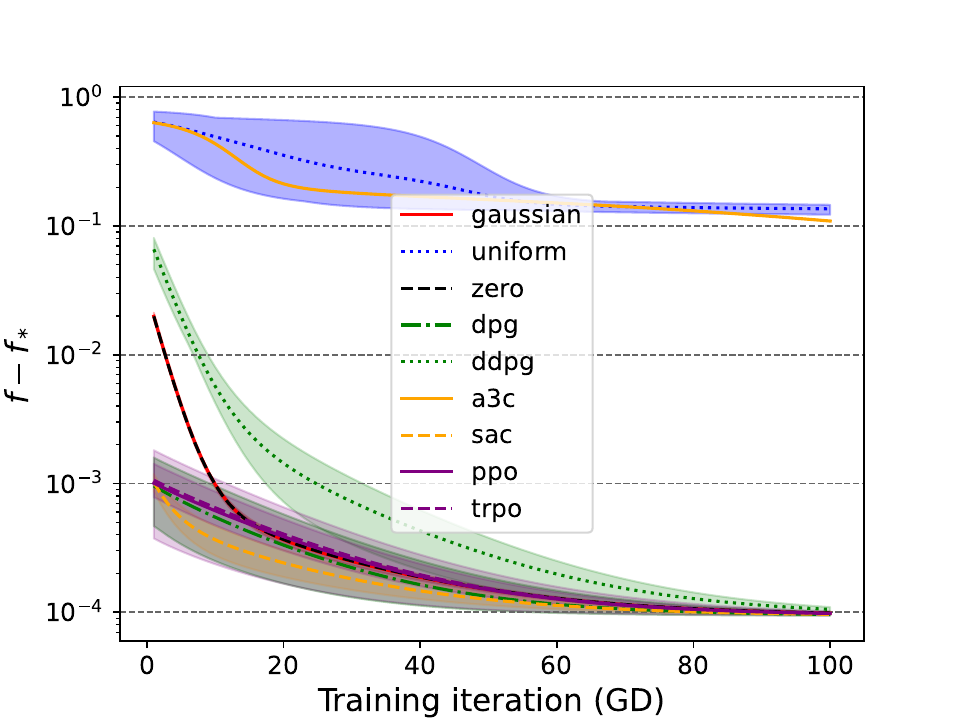}
        \caption{}
        \label{chemenergyfigrl:sub2}
    \end{subfigure}
    \hfill
    \begin{subfigure}[b]{0.3\textwidth}
        \centering
        \includegraphics[width=\textwidth]{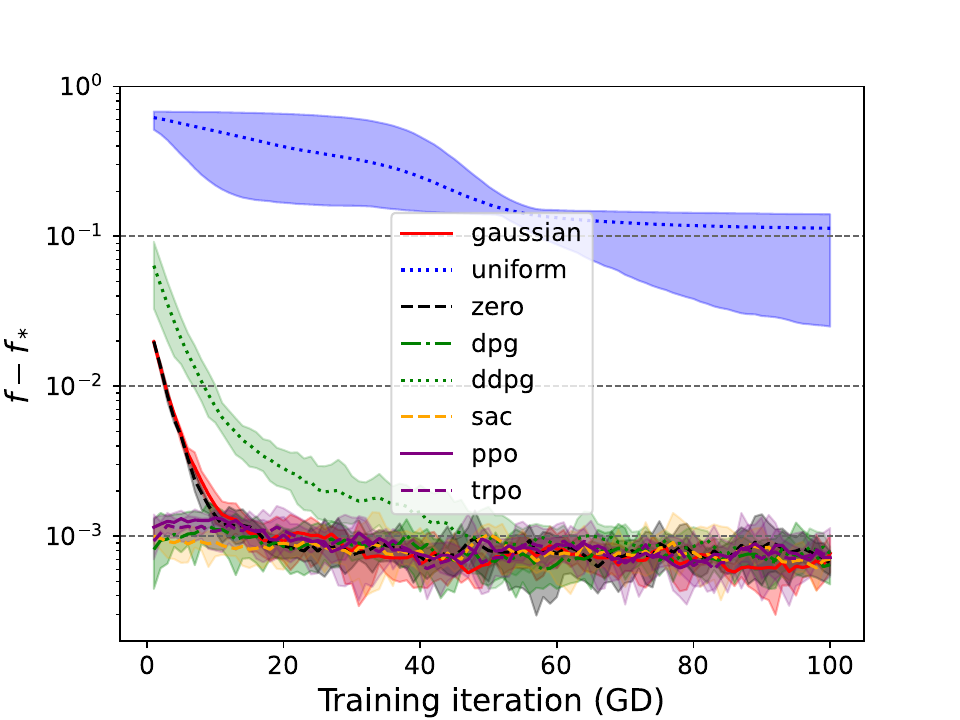}
        \caption{}
        \label{chemenergyfigrl:sub3}
    \end{subfigure}

    \vspace{1em}  

    \begin{subfigure}[b]{0.3\textwidth}
        \centering
        \includegraphics[width=\textwidth]{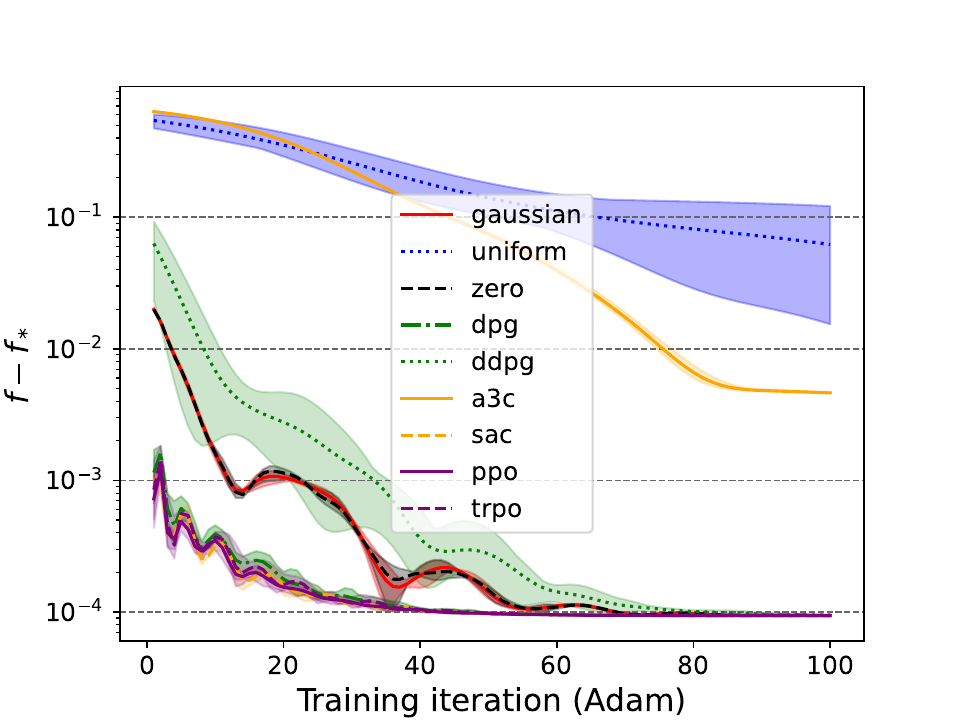}
        \caption{}
        \label{chemenergyfigrl:sub4}
    \end{subfigure}
    \hfill
    \begin{subfigure}[b]{0.3\textwidth}
        \centering
        \includegraphics[width=\textwidth]{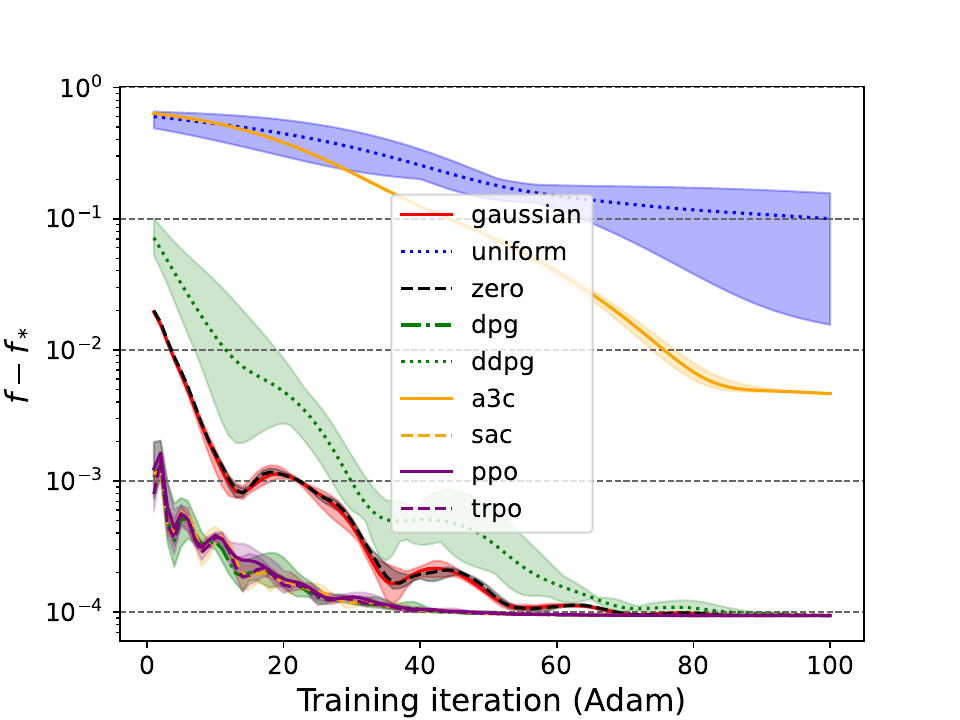}
        \caption{}
        \label{chemenergyfigrl:sub5}
    \end{subfigure}
    \hfill
    \begin{subfigure}[b]{0.3\textwidth}
        \centering
        \includegraphics[width=\textwidth]{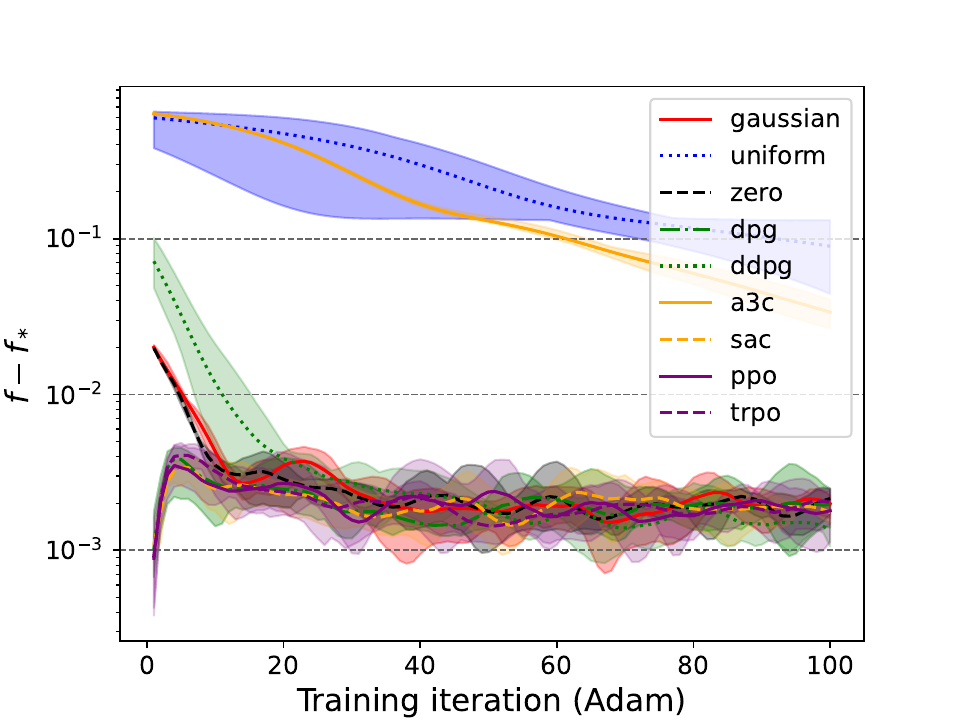}
        \caption{}
        \label{chemenergyfigrl:sub6}
    \end{subfigure}
\caption{Numerical results of loss $f$ minus global minimum $f_\ast$ of the molecule LiH. The first and second rows show training results with the gradient descent and the Adam optimizer, respectively. The left, the middle, and the right columns show results using accurate gradients, noisy gradients with adaptive-distributed noises, and noisy gradients with constant-distributed noises. The variance of noises in the middle line (Figures \ref{chemenergyfigrl:sub2} and \ref{chemenergyfigrl:sub5}) follows Eq. (\ref{eq12}), while the variance of noises in the right line (Figures \ref{chemenergyfigrl:sub3} and \ref{chemenergyfigrl:sub6}) is $0.001$. Each line denotes the average of 5 rounds.}
    \label{fig:all_chemrl}
    \vspace{-2mm}
\end{figure*}

The results in Fig.~\ref{fig:figRLbench} suggest that initializing model parameters via RL can generally outperform traditional random-based initializations (e.g., Gaussian, zero, and uniform baselines) in terms of both convergence speed and final optimization outcome. In particular, DPG demonstrates especially strong performance, as evidenced by the lower gradient norms in Figs.~\ref{figRLbench:sub1}--\ref{figRLbench:sub4}. To investigate further, we isolate DPG for a more granular comparison against a variety of commonly employed initialization methods, including Gaussian, uniform, and zero initialization (see Fig.~\ref{fig:all_five}). Empirically, DPG exhibits a marked advantage, enabling the optimizer to guide the model rapidly toward a region with reduced gradients and to maintain improved convergence throughout the final training epochs (80--100); it is evident that using DPG for parameter initialization expedites convergence to a stable solution, leading to faster optimization compared to conventional methods, while also exhibiting lower variance during training and thus indicating a more robust and reliable convergence behavior.

Despite these advantages, not all RL-based initialization techniques surpass the Gaussian baseline. For example, A3C and DDPG sometimes show less favorable performance, even revealing anomalous increases at later training stages. This indicates that deeper exploration of hyperparameter tuning and algorithmic design is warranted for certain deep RL approaches, such as DDPG. In this work, we have deliberately adopted a single-step training paradigm for simplicity. However, advanced mechanisms such as replay buffers could be incorporated for more stable or robust initialization in future studies.

A similar pattern emerges in Figs.~\ref{figRLbench:sub5}--\ref{figRLbench:sub8}: RL-based initializations do not uniformly dominate, and effectiveness depends heavily on the suitability of each algorithm for the given task. For instance, DDPG often demands more fine-grained parameter tuning, whereas DPG proves simpler and more efficient under identical training conditions. Notably, we employ the same hyperparameter settings and training iterations for all RL algorithms in order to maintain fairness in direct comparisons; however, this uniform treatment may be somewhat disadvantageous for complex deep RL models (PPO, DDPG, TRPO, etc.). Nonetheless, the results underline the potential benefits of DPG methods when constrained to a single-step, homogeneous training regime.

\begin{figure*}
    \centering
  
    \begin{subfigure}[b]{0.3\textwidth}
        \centering
        \includegraphics[width=\textwidth]{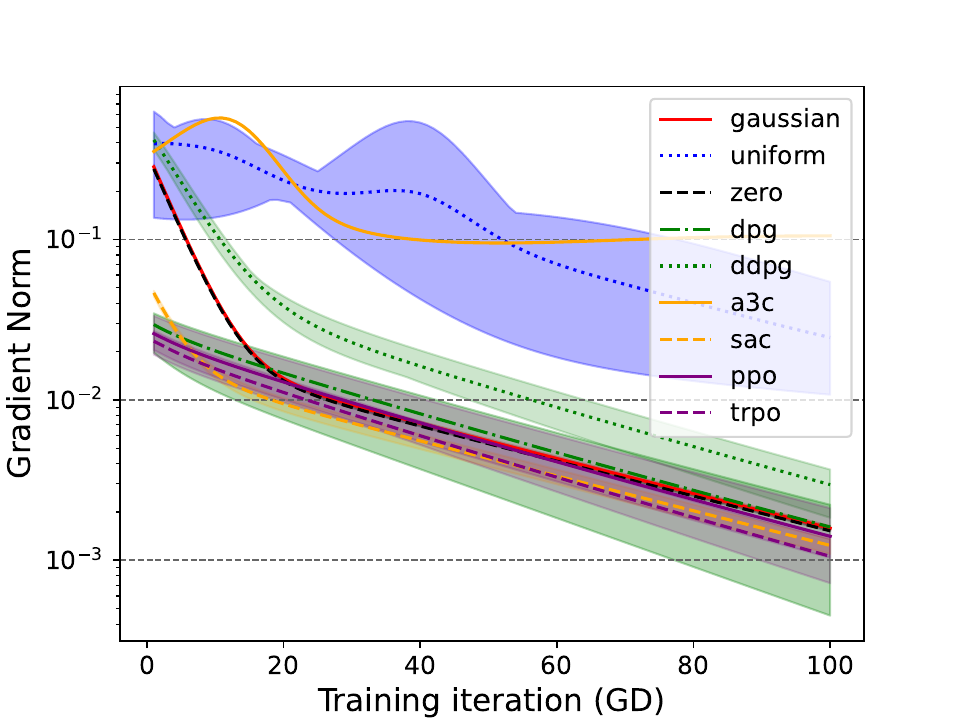}
        \caption{}
        \label{chemnormfigrl:sub1}
    \end{subfigure}
    \hfill
    \begin{subfigure}[b]{0.3\textwidth}
        \centering
        \includegraphics[width=\textwidth]{Figureschem_RL/chem_gradnorm_gd_10_24_0_0.pdf}
        \caption{}
        \label{chemnormfigrl:sub2}
    \end{subfigure}
    \hfill
    \begin{subfigure}[b]{0.3\textwidth}
        \centering
        \includegraphics[width=\textwidth]{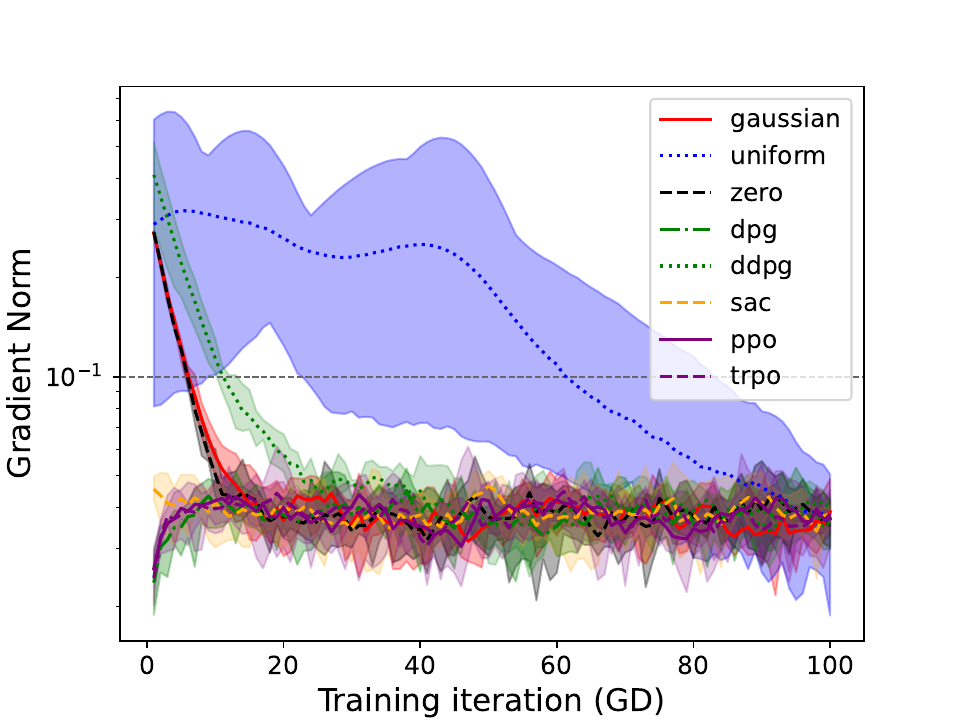}
        \caption{}
        \label{chemnormfigrl:sub3}
    \end{subfigure}

    \vspace{1em}  

    \begin{subfigure}[b]{0.3\textwidth}
        \centering
        \includegraphics[width=\textwidth]{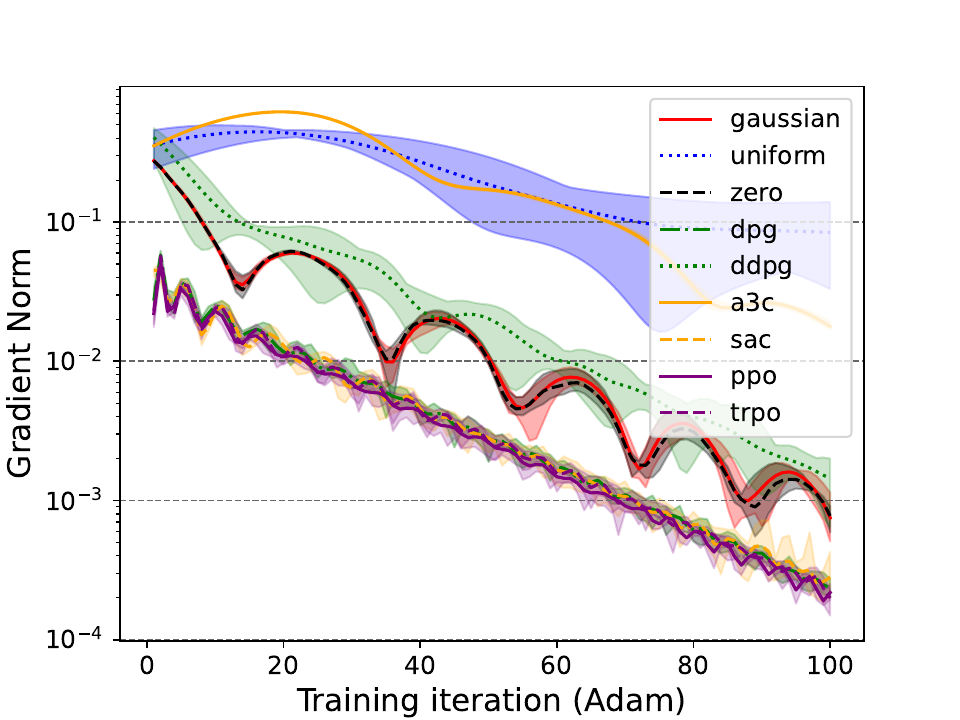}
        \caption{}
        \label{chemnormfigrl:sub4}
    \end{subfigure}
    \hfill
    \begin{subfigure}[b]{0.3\textwidth}
        \centering
        \includegraphics[width=\textwidth]{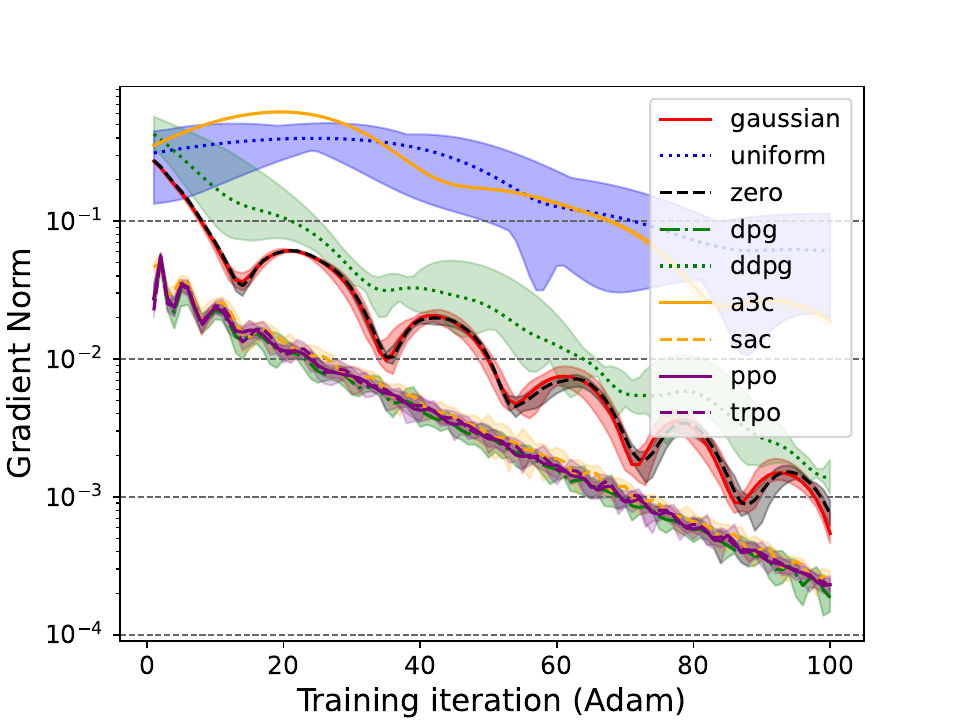}
        \caption{}
        \label{chemnormfigrl:sub5}
    \end{subfigure}
    \hfill
    \begin{subfigure}[b]{0.3\textwidth}
        \centering
        \includegraphics[width=\textwidth]{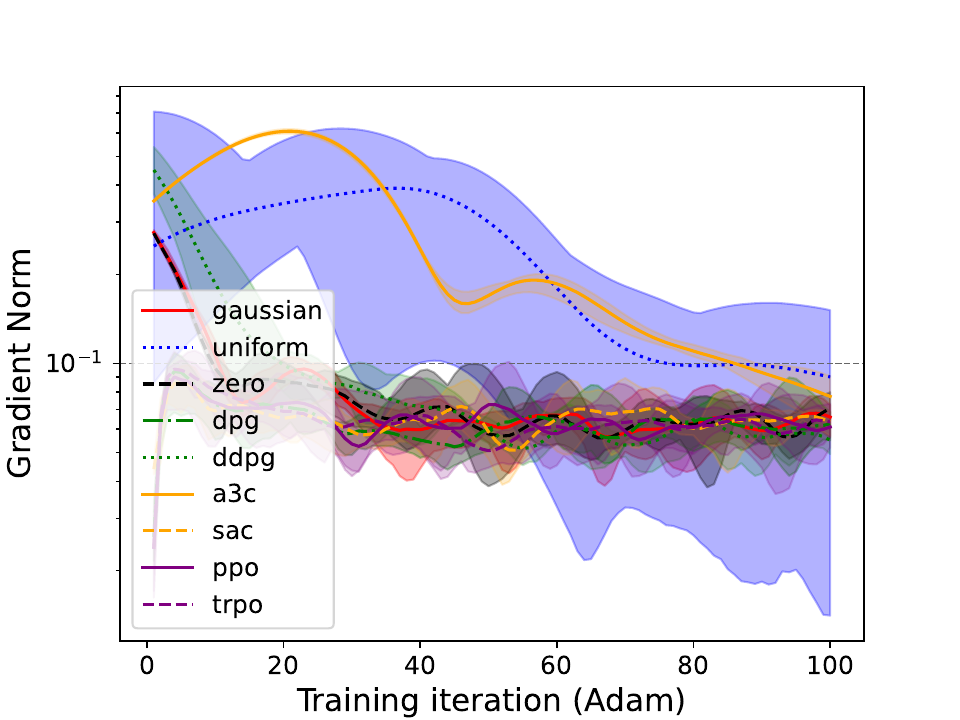}
        \caption{}
        \label{chemnormfigrl:sub6}
    \end{subfigure}

    \caption{Numerical results of finding the ground energy of the molecule LiH. The first and second rows show training results with the gradient descent and the Adam optimizer, respectively. The left, the middle, and the right columns show results using accurate gradients, noisy gradients with
adaptive-distributed noises, and noisy gradients with constant-distributed noises. The variance of noises in the middle line (Figures \ref{chemnormfigrl:sub2} and \ref{chemnormfigrl:sub5}) follows Eq. (\ref{eq12}), while the variance of noises in the right line (Figures \ref{chemnormfigrl:sub3} and \ref{chemnormfigrl:sub6}) is $0.001$. Each line denotes the average of 5 rounds of optimizations.}
    \label{fig:all_chemnormrl}
    \vspace{-2mm}
\end{figure*}

\subsection{LiH Molecule}

In this second numerical experiment, we approximate the ground state and ground energy of the Lithium Hydride (LiH) molecule via a variational quantum algorithm utilizing the Hartree--Fock (HF) reference state and Givens rotation gates, following the setup in \cite{zhang2022escaping}. Concretely, we choose the number of active electrons to be \(n_e = 2\) and the number of free spin orbitals to be \(n_0 = 10\), thus working with \(N = n_0\) qubits. The HF reference state takes the form
\[
  \ket{\phi_{\mathrm{HF}}} 
  \;=\; 
  \Bigl(\ket{1}\Bigr)^{\otimes n_e}
  \;\otimes\; 
  \Bigl(\ket{0}\Bigr)^{\otimes (n_0 - n_e)},
\]
and we construct the parameterized unitary
\[
  V_{\mathrm{Givens}}(\boldsymbol{\theta}) 
  \;=\; 
  \prod_{i=1}^{24} 
  R_{\mathrm{Givens}}^{(i)}(\theta_i),
\]
comprising \(L = 24\) distinct Givens rotation gates, each acting on either two or four qubits.

Following \cite{delgado2021variational}, the LiH Hamiltonian is denoted \(H_{\mathrm{LiH}}\). The associated loss function is
\begin{equation}
\label{eq11}
f(\boldsymbol{\theta})
\;=\;
\mathrm{Tr}\!\Bigl[
  H_{\mathrm{LiH}}\,
  V_{\mathrm{Givens}}(\boldsymbol{\theta})\,
  \ket{\phi_{\mathrm{HF}}}\bra{\phi_{\mathrm{HF}}}\,
  V_{\mathrm{Givens}}(\boldsymbol{\theta})^\dagger
\Bigr],
\end{equation}
and minimizing this expression yields an estimate of the LiH ground energy.

Each Givens rotation acting on two qubits is characterized by \(\bigl(h,a\bigr) = (2,2)\), and each acting on four qubits is labeled \(\bigl(h,a\bigr) = (8,8)\). Consequently, following \cite{zhang2022escaping}, we set the variance $
\gamma^2 
  \;=\; 
  \frac{8^3 \times \tfrac{1}{2}}{48 \times 8^2 \times 24},
$
which aligns with the parameter choice \(\bigl(L,h,a,e\bigr) = (24,8,8,\tfrac12)\). 
As in the Heisenberg model experiment, we consider both exact gradient evaluations and gradients corrupted by fixed-variance noise (\(\sigma^2 = 0.001\)). 
In addition, we investigate a noisy scenario with adaptive noise, where, following \cite{zhang2022escaping}, the noise variance on each partial derivative at iteration \(t\) is given by
\begin{equation}
\label{eq12}
\gamma^2
\;=\;
\frac{1}{96 \,\times\, 24 \,\times\, 8^2}
\;\|\!H_{\mathrm{LiH}}\|\!^2 
\;\left(\frac{\partial f}{\partial \theta}\right)^{2}
\Bigl|_{\theta=\theta^{(t)}},
\end{equation}
ensuring that the noise intensity adapts to the current gradient's magnitude. The learning rates are set to \(0.1\) for GD and \(0.01\) for the Adam optimizer. 

From Figs.~\ref{chemenergyfigrl:sub1}--\ref{chemenergyfigrl:sub2} and Figs.~\ref{chemnormfigrl:sub1}--\ref{chemnormfigrl:sub2}, we observe that, in contrast to the Heisenberg model, DPG no longer retains a pronounced performance advantage for this particular task. Among the evaluated RL-based initialization methods, TRPO demonstrates the most favorable convergence profile, whereas A3C exhibits a distinctly erratic convergence pattern. Although certain RL-based trajectories appear similar to conventional initialization strategies, they still confer noteworthy benefits in terms of both convergence speed and final training outcomes.

These advantages become especially evident when employing the Adam optimizer, as shown in Figs.~\ref{chemenergyfigrl:sub4}-\ref{chemenergyfigrl:sub5} and Figs.~\ref{chemnormfigrl:sub4}--\ref{chemnormfigrl:sub5}. In these figures, RL-initialized approaches 
display a more pronounced and sustained descent in the loss curve, clearly diverging from the trends associated with Gaussian, uniform, or zero initializations. Such distinctions underscore the potential of RL-based initialization to more effectively traverse complex parameter landscapes, even if certain algorithms (e.g., A3C) display atypical behavior throughout the optimization trajectory. This outcome highlights the nuanced interplay between algorithmic selection, model-specific structure, and hyperparameter configurations in achieving robust and efficient convergence.

\section{Conclusion}
\label{sec:conclusions}
This paper proposes an initialization strategy based on RL to effectively alleviate the common Barren Plateau problem in parameter training in VQA and verifies its feasibility and superiority through various experiments. Compared with traditional initialization (such as Gaussian distribution, zero initialization, or uniform distribution), the RL initialization method proposed in this paper can help parameters stay away from the area where the gradient is extremely small or even asymptotically zero in the early stage of optimization, thereby significantly reducing the gradient vanishing phenomenon caused by deep quantum circuits. Through numerical simulations under different noise conditions and different types of tasks, compared with traditional methods such as random initialization, RL initialization can find a better parameter distribution in the early stage of optimization so that subsequent classical optimization methods such as gradient descent or Adam can converge to a lower objective function value in a shorter time. It provides new ideas for the research and practical deployment of variational quantum algorithms on a larger scale in the future and also proves that interdisciplinary integration can produce important synergistic effects at the algorithm level. In the future, we can further explore more complex quantum systems, more realistic experimental noise models, and RL designs under multiple incentives, providing more theoretical and experimental support for the scalability and practicality of variational quantum algorithms.

\clearpage

\newpage

\bibliographystyle{IEEEtran}
\bibliography{reference}

\begin{thebibliography}{10}
\providecommand{\url}[1]{#1}
\csname url@samestyle\endcsname
\providecommand{\newblock}{\relax}
\providecommand{\bibinfo}[2]{#2}
\providecommand{\BIBentrySTDinterwordspacing}{\spaceskip=0pt\relax}
\providecommand{\BIBentryALTinterwordstretchfactor}{4}
\providecommand{\BIBentryALTinterwordspacing}{\spaceskip=\fontdimen2\font plus
\BIBentryALTinterwordstretchfactor\fontdimen3\font minus \fontdimen4\font\relax}
\providecommand{\BIBforeignlanguage}[2]{{%
\expandafter\ifx\csname l@#1\endcsname\relax
\typeout{** WARNING: IEEEtran.bst: No hyphenation pattern has been}%
\typeout{** loaded for the language `#1'. Using the pattern for}%
\typeout{** the default language instead.}%
\else
\language=\csname l@#1\endcsname
\fi
#2}}
\providecommand{\BIBdecl}{\relax}
\BIBdecl

\bibitem{cerezo2021variational}
M.~Cerezo, A.~Arrasmith, R.~Babbush, S.~C. Benjamin, S.~Endo, K.~Fujii, J.~R. McClean, K.~Mitarai, X.~Yuan, L.~Cincio \emph{et~al.}, ``Variational quantum algorithms,'' \emph{Nature Reviews Physics}, vol.~3, no.~9, pp. 625--644, 2021.

\bibitem{bharti2022noisy}
K.~Bharti, A.~Cervera-Lierta, T.~H. Kyaw, T.~Haug, S.~Alperin-Lea, A.~Anand, M.~Degroote, H.~Heimonen, J.~S. Kottmann, T.~Menke \emph{et~al.}, ``Noisy intermediate-scale quantum algorithms,'' \emph{Reviews of Modern Physics}, vol.~94, no.~1, p. 015004, 2022.

\bibitem{schuld2020circuit}
M.~Schuld, A.~Bocharov, K.~M. Svore, and N.~Wiebe, ``Circuit-centric quantum classifiers,'' \emph{Physical Review A}, vol. 101, no.~3, p. 032308, 2020.

\bibitem{mitarai2018quantum}
K.~Mitarai, M.~Negoro, M.~Kitagawa, and K.~Fujii, ``Quantum circuit learning,'' \emph{Physical Review A}, vol.~98, no.~3, p. 032309, 2018.

\bibitem{chen2022quantumLSTM}
S.~Y.-C. Chen, S.~Yoo, and Y.-L.~L. Fang, ``Quantum long short-term memory,'' in \emph{ICASSP 2022-2022 IEEE International Conference on Acoustics, Speech and Signal Processing (ICASSP)}.\hskip 1em plus 0.5em minus 0.4em\relax IEEE, 2022, pp. 8622--8626.

\bibitem{di2022dawn}
R.~Di~Sipio, J.-H. Huang, S.~Y.-C. Chen, S.~Mangini, and M.~Worring, ``The dawn of quantum natural language processing,'' in \emph{ICASSP 2022-2022 IEEE International Conference on Acoustics, Speech and Signal Processing (ICASSP)}.\hskip 1em plus 0.5em minus 0.4em\relax IEEE, 2022, pp. 8612--8616.

\bibitem{stein2023applying}
J.~Stein, I.~Christ, N.~Kraus, M.~B. Mansky, R.~M{\"u}ller, and C.~Linnhoff-Popien, ``Applying qnlp to sentiment analysis in finance,'' in \emph{2023 IEEE International Conference on Quantum Computing and Engineering (QCE)}, vol.~2.\hskip 1em plus 0.5em minus 0.4em\relax IEEE, 2023, pp. 20--25.

\bibitem{li2023pqlm}
S.~S. Li, X.~Zhang, S.~Zhou, H.~Shu, R.~Liang, H.~Liu, and L.~P. Garcia, ``Pqlm-multilingual decentralized portable quantum language model,'' in \emph{ICASSP 2023-2023 IEEE International Conference on Acoustics, Speech and Signal Processing (ICASSP)}.\hskip 1em plus 0.5em minus 0.4em\relax IEEE, 2023, pp. 1--5.

\bibitem{chen2022quantumCNN}
S.~Y.-C. Chen, T.-C. Wei, C.~Zhang, H.~Yu, and S.~Yoo, ``Quantum convolutional neural networks for high energy physics data analysis,'' \emph{Physical Review Research}, vol.~4, no.~1, p. 013231, 2022.

\bibitem{yun2023quantum}
W.~J. Yun, J.~Park, and J.~Kim, ``Quantum multi-agent meta reinforcement learning,'' in \emph{Proceedings of the AAAI Conference on Artificial Intelligence}, vol.~37, no.~9, 2023, pp. 11\,087--11\,095.

\bibitem{lockwood2020reinforcement}
O.~Lockwood and M.~Si, ``Reinforcement learning with quantum variational circuit,'' in \emph{Proceedings of the AAAI conference on artificial intelligence and interactive digital entertainment}, vol.~16, no.~1, 2020, pp. 245--251.

\bibitem{skolik2022quantum}
A.~Skolik, S.~Jerbi, and V.~Dunjko, ``Quantum agents in the gym: a variational quantum algorithm for deep q-learning,'' \emph{Quantum}, vol.~6, p. 720, 2022.

\bibitem{jerbi2021parametrized}
S.~Jerbi, C.~Gyurik, S.~Marshall, H.~Briegel, and V.~Dunjko, ``Parametrized quantum policies for reinforcement learning,'' \emph{Advances in Neural Information Processing Systems}, vol.~34, pp. 28\,362--28\,375, 2021.

\bibitem{chen2020variational}
S.~Y.-C. Chen, C.-H.~H. Yang, J.~Qi, P.-Y. Chen, X.~Ma, and H.-S. Goan, ``Variational quantum circuits for deep reinforcement learning,'' \emph{IEEE access}, vol.~8, pp. 141\,007--141\,024, 2020.

\bibitem{peng2024hyq2}
Y.~Peng, X.~Li, Z.~Liang, and Y.~Wang, ``Hyq2: A hybrid quantum neural network for nextg vulnerability detection,'' \emph{IEEE Transactions on Quantum Engineering}, 2024.

\bibitem{peng2024qrng}
Y.~Peng, X.~Li, and Y.~Wang, ``Qrng-ddpm: Enhancing diffusion models through fitting mixture noise with quantum random number,'' in \emph{2024 IEEE International Conference on Quantum Computing and Engineering (QCE)}, vol.~2.\hskip 1em plus 0.5em minus 0.4em\relax IEEE, 2024, pp. 92--96.

\bibitem{peng2024qsco}
Y.~Peng, X.~Li, Z.~Liang, and Y.~Wang, ``Qsco: A quantum scoring module for open-set supervised anomaly detection,'' \emph{arXiv preprint arXiv:2405.16368}, 2024.

\bibitem{peng2024quantum}
Y.~Peng, X.~Li, and Y.~Wang, ``Quantum squeeze-and-excitation networks,'' in \emph{2024 IEEE International Conference on Quantum Computing and Engineering (QCE)}, vol.~2.\hskip 1em plus 0.5em minus 0.4em\relax IEEE, 2024, pp. 39--43.

\bibitem{peng2024hybrid}
Y.~Peng, X.~Li, Z.~Liang, and Y.~Wang, ``Hybrid quantum downsampling networks,'' \emph{arXiv preprint arXiv:2405.16375}, 2024.

\bibitem{peng2025quantum}
Y.~Peng, D.~Li, X.~Li, Z.~Liang, Y.~Ding, and Y.~Wang, ``Quantum-inspired fidelity-based divergence,'' \emph{arXiv preprint arXiv:2501.19307}, 2025.

\bibitem{mcclean2018barren}
J.~R. McClean, S.~Boixo, V.~N. Smelyanskiy, R.~Babbush, and H.~Neven, ``Barren plateaus in quantum neural network training landscapes,'' \emph{Nature communications}, vol.~9, no.~1, p. 4812, 2018.

\bibitem{cerezo2021cost}
M.~Cerezo, A.~Sone, T.~Volkoff, L.~Cincio, and P.~J. Coles, ``Cost function dependent barren plateaus in shallow parametrized quantum circuits,'' \emph{Nature communications}, vol.~12, no.~1, p. 1791, 2021.

\bibitem{pesah2021absence}
A.~Pesah, M.~Cerezo, S.~Wang, T.~Volkoff, A.~T. Sornborger, and P.~J. Coles, ``Absence of barren plateaus in quantum convolutional neural networks,'' \emph{Physical Review X}, vol.~11, no.~4, p. 041011, 2021.

\bibitem{larocca2024review}
M.~Larocca, S.~Thanasilp, S.~Wang, K.~Sharma, J.~Biamonte, P.~J. Coles, L.~Cincio, J.~R. McClean, Z.~Holmes, and M.~Cerezo, ``A review of barren plateaus in variational quantum computing,'' \emph{arXiv preprint arXiv:2405.00781}, 2024.

\bibitem{sutton1998reinforcement}
R.~S. Sutton, A.~G. Barto \emph{et~al.}, \emph{Reinforcement learning: An introduction}.\hskip 1em plus 0.5em minus 0.4em\relax MIT press Cambridge, 1998, vol.~1, no.~1.

\bibitem{silver2014deterministic}
D.~Silver, G.~Lever, N.~Heess, T.~Degris, D.~Wierstra, and M.~Riedmiller, ``Deterministic policy gradient algorithms,'' in \emph{International conference on machine learning}.\hskip 1em plus 0.5em minus 0.4em\relax Pmlr, 2014, pp. 387--395.

\bibitem{wen2022multi}
M.~Wen, J.~Kuba, R.~Lin, W.~Zhang, Y.~Wen, J.~Wang, and Y.~Yang, ``Multi-agent reinforcement learning is a sequence modeling problem,'' \emph{Advances in Neural Information Processing Systems}, vol.~35, pp. 16\,509--16\,521, 2022.

\bibitem{fujimoto2021minimalist}
S.~Fujimoto and S.~S. Gu, ``A minimalist approach to offline reinforcement learning,'' \emph{Advances in neural information processing systems}, vol.~34, pp. 20\,132--20\,145, 2021.

\bibitem{salimans2017evolution}
T.~Salimans, J.~Ho, X.~Chen, S.~Sidor, and I.~Sutskever, ``Evolution strategies as a scalable alternative to reinforcement learning,'' \emph{arXiv preprint arXiv:1703.03864}, 2017.

\bibitem{DPG}
D.~Silver, G.~Lever, N.~Heess, T.~Degris, D.~Wierstra, and M.~Riedmiller, ``Deterministic policy gradient algorithms,'' in \emph{International conference on machine learning}.\hskip 1em plus 0.5em minus 0.4em\relax Pmlr, 2014, pp. 387--395.

\bibitem{TRPO}
J.~Schulman, S.~Levine, P.~Abbeel, M.~Jordan, and P.~Moritz, ``Trust region policy optimization,'' in \emph{International conference on machine learning}.\hskip 1em plus 0.5em minus 0.4em\relax PMLR, 2015, pp. 1889--1897.

\bibitem{A3C}
V.~Mnih, A.~P. Badia, M.~Mirza, A.~Graves, T.~Lillicrap, T.~Harley, D.~Silver, and K.~Kavukcuoglu, ``Asynchronous methods for deep reinforcement learning,'' in \emph{International conference on machine learning}.\hskip 1em plus 0.5em minus 0.4em\relax PmLR, 2016, pp. 1928--1937.

\bibitem{PPO}
J.~Schulman, F.~Wolski, P.~Dhariwal, A.~Radford, and O.~Klimov, ``Proximal policy optimization algorithms,'' \emph{arXiv preprint arXiv:1707.06347}, 2017.

\bibitem{DDPG}
T.~P. Lillicrap, J.~J. Hunt, A.~Pritzel, N.~Heess, T.~Erez, Y.~Tassa, D.~Silver, and D.~Wierstra, ``Continuous control with deep reinforcement learning,'' \emph{arXiv preprint arXiv:1509.02971}, 2015.

\bibitem{SAC}
T.~Haarnoja, A.~Zhou, K.~Hartikainen, G.~Tucker, S.~Ha, J.~Tan, V.~Kumar, H.~Zhu, A.~Gupta, P.~Abbeel \emph{et~al.}, ``Soft actor-critic algorithms and applications,'' \emph{arXiv preprint arXiv:1812.05905}, 2018.

\bibitem{grant2019initialization}
E.~Grant, L.~Wossnig, M.~Ostaszewski, and M.~Benedetti, ``An initialization strategy for addressing barren plateaus in parametrized quantum circuits,'' \emph{Quantum}, vol.~3, p. 214, 2019.

\bibitem{zhang2022escaping}
K.~Zhang, L.~Liu, M.-H. Hsieh, and D.~Tao, ``Escaping from the barren plateau via gaussian initializations in deep variational quantum circuits,'' \emph{Advances in Neural Information Processing Systems}, vol.~35, pp. 18\,612--18\,627, 2022.

\bibitem{volkoff2021large}
T.~Volkoff and P.~J. Coles, ``Large gradients via correlation in random parameterized quantum circuits,'' \emph{Quantum Science and Technology}, vol.~6, no.~2, p. 025008, 2021.

\bibitem{skolik2021layerwise}
A.~Skolik, J.~R. McClean, M.~Mohseni, P.~Van Der~Smagt, and M.~Leib, ``Layerwise learning for quantum neural networks,'' \emph{Quantum Machine Intelligence}, vol.~3, pp. 1--11, 2021.

\bibitem{mari2020transfer}
A.~Mari, T.~R. Bromley, J.~Izaac, M.~Schuld, and N.~Killoran, ``Transfer learning in hybrid classical-quantum neural networks,'' \emph{Quantum}, vol.~4, p. 340, 2020.

\bibitem{henderson2020quanvolutional}
M.~Henderson, S.~Shakya, S.~Pradhan, and T.~Cook, ``Quanvolutional neural networks: powering image recognition with quantum circuits,'' \emph{Quantum Machine Intelligence}, vol.~2, no.~1, p.~2, 2020.

\bibitem{rivera2021avoiding}
J.~Rivera-Dean, P.~Huembeli, A.~Ac{\'\i}n, and J.~Bowles, ``Avoiding local minima in variational quantum algorithms with neural networks,'' \emph{arXiv preprint arXiv:2104.02955}, 2021.

\bibitem{akshay2020reachability}
V.~Akshay, H.~Philathong, M.~E. Morales, and J.~D. Biamonte, ``Reachability deficits in quantum approximate optimization,'' \emph{Physical review letters}, vol. 124, no.~9, p. 090504, 2020.

\bibitem{larocca2022diagnosing}
M.~Larocca, P.~Czarnik, K.~Sharma, G.~Muraleedharan, P.~J. Coles, and M.~Cerezo, ``Diagnosing barren plateaus with tools from quantum optimal control,'' \emph{Quantum}, vol.~6, p. 824, 2022.

\bibitem{ragone2024lie}
M.~Ragone, B.~N. Bakalov, F.~Sauvage, A.~F. Kemper, C.~Ortiz~Marrero, M.~Larocca, and M.~Cerezo, ``A lie algebraic theory of barren plateaus for deep parameterized quantum circuits,'' \emph{Nature Communications}, vol.~15, no.~1, p. 7172, 2024.

\bibitem{schulman2015trust}
J.~Schulman, S.~Levine, P.~Abbeel, M.~Jordan, and P.~Moritz, ``Trust region policy optimization,'' in \emph{International conference on machine learning}.\hskip 1em plus 0.5em minus 0.4em\relax PMLR, 2015, pp. 1889--1897.

\bibitem{bonechi1992heisenberg}
F.~Bonechi, E.~Celeghini, R.~Giachetti, E.~Sorace, and M.~Tarlini, ``Heisenberg xxz model and quantum galilei group,'' \emph{Journal of Physics A: Mathematical and General}, vol.~25, no.~15, p. L939, 1992.

\bibitem{delgado2021variational}
A.~Delgado, J.~M. Arrazola, S.~Jahangiri, Z.~Niu, J.~Izaac, C.~Roberts, and N.~Killoran, ``Variational quantum algorithm for molecular geometry optimization,'' \emph{Physical Review A}, vol. 104, no.~5, p. 052402, 2021.

\end{thebibliography}

\end{document}